%% file: neurips_2026.tex
\documentclass{article}

\usepackage{graphicx}
 \usepackage[preprint]{neurips_2026}

\makeatletter
\renewcommand{\@noticestring}{%
  ICML 2026 Workshop on AdaptFM: Resource-Adaptive Foundation Model Inference.%
}
\makeatother

\usepackage[utf8]{inputenc} 
\usepackage[T1]{fontenc}    
\usepackage{hyperref}       
\usepackage{url}            
\usepackage{booktabs}       
\usepackage{amsfonts}       
\usepackage{nicefrac}       
\usepackage{microtype}      
\usepackage{xcolor}         
\usepackage{natbib}
\usepackage{amsmath}

\usepackage{cleveref}
\usepackage{multirow}
\usepackage{float}

\title{You Had One Job: Per-Task Quantization Using LLMs’ Hidden Representations}

\author{%
  \textbf{Amit LeVi}\textsuperscript{1,*} \quad
  \textbf{Raz Lapid}\textsuperscript{2,*} \quad
  \textbf{Rom Himelstein}\textsuperscript{1} \quad
  \textbf{Chaim Baskin}\textsuperscript{3} \\
  \textbf{Ravid Shwartz-Ziv}\textsuperscript{4} \quad
  \textbf{Avi Mendelson}\textsuperscript{1} \\[0.5em]
  \textsuperscript{1}Technion---Israel Institute of Technology \quad
  \textsuperscript{2}Intuit \\
  \textsuperscript{3}Ben-Gurion University of the Negev \quad
  \textsuperscript{4}New York University \\
  \textsuperscript{*}Equal contribution
}

\begin{document}

\maketitle
\input{sec/0_abstract}

\input{sec/1_intro}

\input{sec/2_related_work}
\input{sec/3_motivation}

\input{sec/4_method}
\input{sec/5_experiments}

\input{sec/6_related_work}
\input{sec/7_conclusions}

\bibliographystyle{plainnat} 
\bibliography{neurips_2026}

\appendix

\input{sec/8_appendix}

\newpage
\input{checklist.tex}

\end{document}

%% file: sec/0_abstract.tex
\begin{abstract}
Many LLM applications require only narrow capabilities, yet standard post-training quantization (PTQ) methods allocate precision without considering the target task. This can waste bits on layers that are less relevant to the task signal while over-compressing layers that are critical for downstream behavior. We propose Task-Aware Quantization (TAQ), a training-free, weight-only mixed-precision PTQ framework that uses a small set of unlabeled task calibration prompts to allocate higher precision to task-relevant transformer layers under a fixed bit budget. TAQ estimates layer importance from hidden representations and output sensitivity, and we instantiate it with three scoring rules: TAQ-IS, based on activation information and stability; TAQ-KL, based on output-distribution sensitivity under a quantization-noise proxy; and TAQ-O, a label-informed oracle diagnostic for analyzing layer sensitivity. Across several benchmarks, TAQ outperforms task-agnostic baselines such in most settings, with especially strong gains in the accuracy--memory ratio. We further validate that these gains translate to real deployment behavior through hardware throughput and latency measurements, and analyze calibration robustness and residual-stream error propagation. Overall, TAQ turns mixed-precision PTQ from a model-centric compression step into a task-conditioned precision-allocation problem. A reference implementation is available at \href{https://anonymous.4open.science/r/TAQ-9217/README.md}{\includegraphics[height=1em]{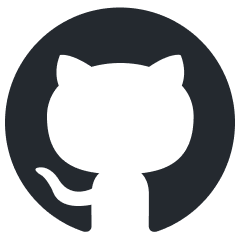}}.
\end{abstract}

%% file: sec/1_intro.tex
\section{Introduction}
\label{section:intro}

Large language models (LLMs) continue to scale in size and capability \citep{vaswani2017attention,kaplan2020scaling}, but this scaling substantially increases inference and memory costs \citep{pope2023efficiently}. At the same time, many deployed applications require only a narrow subset of model capabilities, such as code completion~\citep{chen2021evaluating,roziere2023code}, mathematical reasoning \citep{cobbe2021training,hendrycks2021measuring}, or domain-specific question answering. This motivates compression methods that can preserve the capabilities most relevant to the target use case while avoiding task-specific training or separate fine-tuned models.

Post-training quantization (PTQ) reduces memory footprint and inference cost by representing model weights or activations with fewer bits \citep{gholami2022survey}. Existing LLM PTQ methods include accurate weight reconstruction methods \citep{frantar2022gptq,guan2024aptq}, activation-aware and outlier-mitigation techniques \citep{dettmers2022gpt3,xiao2023smoothquant}, and mixed-precision schemes based on global or layer-sensitive criteria \citep{dong2019hawq,lin2024awq,zhang2025towards,zhao2025coopq}. However, these methods are largely task-agnostic: they optimize generic reconstruction, activation, curvature, or salience proxies rather than asking which layers are most important for a particular downstream task.

\input{figs/frame}

This mismatch is important because transformer layers contribute heterogeneously to end-task behavior \citep{tenney2019bert,meng2022locating,geva2021transformer}. Work in representation analysis and mechanistic interpretability suggests that task-relevant information is not uniformly distributed across depth, and that hidden states can expose where task-conditioned computation is concentrated \citep{cunningham2023sparse,heimersheim2024use,sharkey2025open}. This motivates a simple question: given a frozen LLM, a small calibration set from a target task, and a memory budget, where should precision be spent? 

In this work, we introduce a novel task-conditioned precision-allocation approach for post-training mixed-precision quantization. We propose \textbf{Task-Aware Quantization} (\textbf{TAQ}), a training-free, weight-only PTQ framework that uses task-specific calibration prompts to estimate layer importance with respect to the residual-stream task signal and allocate precision under a fixed budget, an novel precision-allocation policies that can be combined with existing PTQ backends. We instantiate TAQ with three layer-scoring rules: \textbf{TAQ-IS} (\textbf{I}nformation and \textbf{S}tability), \textbf{TAQ-KL} (\textbf{KL}-divergence-based output sensitivity), and \textbf{TAQ-O} (\textbf{O}racle sensitivity), a label-informed diagnostic used for analysis rather than deployment (Fig.~\ref{fig:framwork}). To the best of our knowledge, TAQ is the first training-free, weight-only mixed-precision PTQ framework to allocate per-layer precision from unlabeled target-task prompts, using residual-stream hidden-state statistics and output-sensitivity signals. Our contributions are:
\begin{itemize}
    \item \textbf{(1) Task-aware PTQ.} We formulate mixed-precision PTQ as task-conditioned precision allocation and introduce TAQ.
    \item \textbf{(2) Layer-importance scoring.} We propose TAQ-IS, TAQ-KL, and diagnostic TAQ-O to identify task-sensitive layers for higher precision.
    \item \textbf{(3) Accuracy--memory--latency gains.} Across code, reasoning, and knowledge tasks, TAQ improves accuracy--memory trade-offs over GPTQ, AWQ, and SliM-LLM, with hardware validation.
    \item \textbf{(4) Robustness analysis.} We study calibration size, mixed-task calibration, structural baselines, and inter-layer effects, showing that TAQ rankings are largely stable in practical settings.
\end{itemize}

%% file: figs/frame.tex
\begin{figure*}
    \centering
    \includegraphics[width=1\linewidth]{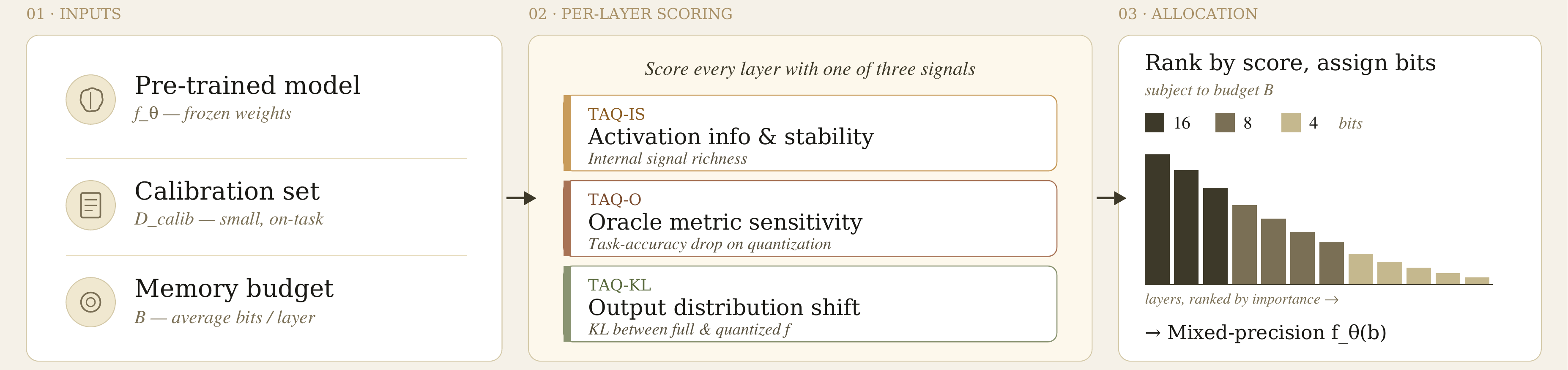}
    \caption{Overview of the framework evaluating layer importance via \emph{TAQ-IS, TAQ-O, TAQ-KL}.}
    \label{fig:framwork}
\end{figure*}

%% file: sec/2_related_work.tex
\section{Related Work}
\label{related}
PTQ reduces LLM memory and inference cost while preserving general behavior through weight reconstruction, outlier handling, and calibration \citep{frantar2022gptq,xiao2023smoothquant,lin2024awq,shao2023omniquant,dettmers2023spqr}. Mixed-precision methods further show that layers, channels, and groups differ in sensitivity and need not use uniform bit-widths, typically allocating precision according to model-centric criteria such as curvature, reconstruction error, salience, or average zero-shot fidelity \citep{dong2019hawq,guan2024aptq,huang2024slim,zhang2025towards,zhao2025coopq}. Since component importance varies and high-importance components can benefit from higher precision \citep{zheng2024mixllm}, these methods provide a natural starting point for adaptive bit allocation; however, they generally do not explicitly ask which layers are most important for a specific downstream task. TAQ instead treats bit allocation as a task-conditioned decision: using a small unlabeled calibration set from the target distribution, it identifies layers that preserve task-relevant residual-stream signal and assigns them higher precision under a fixed bit budget. This is motivated by complementary work in representation analysis and mechanistic interpretability showing that hidden representations can encode semantic, factual, task, or behavioral features, and that localized activation interventions can reveal where task-relevant computations are concentrated or steer model behavior \citep{turner2023steering,zou2023representation,meng2022locating,shnaidman2025activation,rahimi2026step,himelstein2026silenced}. These findings motivate using hidden-state statistics as lightweight diagnostics for where task information is expressed across depth, without requiring exact circuit localization.

\section{Background}
\label{back}
We use this perspective to define task-conditioned layer statistics. Let $D_{\mathrm{calib}}=\{q_j\}_{j=1}^{n}$ be a small calibration set from the target task. For an input prompt $q_j$ with $T_j$ valid tokens, let
$H_\ell(q_j)\in\mathbb{R}^{T_j\times d}$
denote the hidden states produced by layer $\ell$. Aggregating valid token representations from prompts in $D_{\mathrm{calib}}$ gives a layer-wise representation matrix
$X_\ell\in\mathbb{R}^{m_\ell\times d}$,
where $m_\ell=\sum_{j=1}^{n}T_j$ is the number of collected token vectors. To measure how dispersed these representations are, we form the token--token Gram matrix
$G_\ell=X_\ell X_\ell^\top$.
Let $\{\lambda_{\ell,i}\}_{i=1}^{m_\ell}$ denote its nonnegative eigenvalues and
$\tilde{\lambda}_{\ell,i}=\lambda_{\ell,i}/\sum_k \lambda_{\ell,k}$.
The spectral entropy of the layer representation is

\begin{equation}
H(G_\ell)
=
-\sum_{i=1}^{m_\ell}
\tilde{\lambda}_{\ell,i}
\log
\left(
\tilde{\lambda}_{\ell,i}+\epsilon
\right).
\label{eq:entropy}
\end{equation}

Larger entropy indicates that task prompts induce a more distributed representation at layer $\ell$, whereas lower entropy indicates concentration along fewer dominant directions. Together with simple activation statistics such as variance, this provides a label-free way to characterize how task inputs are represented across depth. Since task-useful information is often not uniformly distributed across layers~\citep{tenney2019bert,jawahar2019does,meng2022locating,hendel2023context,skean2025layer}, minimizing task-agnostic quantization distortion may fail to preserve the computations most relevant to a target task. TAQ uses these task-conditioned representation diagnostics to guide per-layer precision allocation.

%% file: sec/3_motivation.tex

%% file: sec/4_method.tex
\section{Per-task quantization}
\label{sec:per_task_quantization}

We first motivate task-aware precision allocation by examining two empirical properties of LLM representations: (i) task prompts induce different layer-wise representation profiles, and (ii) quantization affects hidden representations non-uniformly across depth. These observations motivate the TAQ framework introduced in the following subsection. \\\\
\textbf{Task-conditioned layer profiles.} Prior work suggests that different tasks rely on different regions of an LLM during inference~\citep{skean2025layer}. We examine this effect by computing the spectral entropy of layer-wise hidden representations (\autoref{eq:entropy}) on calibration prompts from several tasks, the entropy profiles vary across tasks and depth, indicating that task inputs induce distinct representational structure across layers. This supports using task-conditioned hidden-state statistics as a signal for precision allocation.

\input{figs/motivation_layers}

\subsection{Layer-wise task-aware quantization framework}
\label{sec:method}

Given a downstream task, let \(\mathcal{D}_{\mathrm{task}}\) denote the unknown task distribution, and let
\(\mathcal{D}_{\mathrm{calib}}=\{x_i\}_{i=1}^{n}\) be a small unlabeled calibration set sampled from its input marginal. Our goal is to allocate precision across layers so as to preserve task performance while satisfying a memory or compute budget.

Let \(f_\theta\) be a language model with \(L\) transformer blocks and hidden size \(d\). Given a task loss \(\mathcal{L}\) or surrogate objective and a layer-wise cost model \(c_\ell(b_\ell)\), we select a per-layer precision assignment \(\mathbf{b}=(b_1,\dots,b_L)\), where \(b_\ell\in\{4,8,16\}\). The resulting quantized model is denoted by \(f_{\theta}^\mathbf{b}\). Formally, we aim to solve:
\begin{equation}
\min_{\mathbf{b}}\quad
\mathbb{E}_{(x,y)\sim\mathcal{D}_{\mathrm{task}}}\!\left[
\mathcal{L}\!\left(f_{\theta}^\mathbf{b}(x),y\right)
\right]
\quad
\text{s.t.}\quad
\sum_{\ell=1}^{L} c_\ell(b_\ell)\le B,
\label{eq:objective}
\end{equation}
where \(B\) denotes the overall budget. Directly solving
Eq.~\eqref{eq:objective} is expensive because it would require evaluating many
candidate bit allocations under the downstream task objective. We therefore
approximate it by estimating task-conditioned layer-importance scores \(s_\ell\)
from the calibration set \(\mathcal{D}_{\mathrm{calib}}\), and allocate higher
precision to layers with larger scores. In our experiments, TAQ-IS and TAQ-KL instantiate this framework with \(b_\ell\in\{4,8\}\), while TAQ-O uses \(b_\ell\in\{4,16\}\).
\paragraph{Quantization operator: weight-only, group-wise affine.}
For each linear weight matrix, we apply weight-only per-group affine quantization with group size \(G\). For a weight group \(w\in\mathbb{R}^{G}\) and bitwidth \(b\in\{4,8\}\), let \(Q_b=2^b-1\). The integer quantized weights and their dequantized approximation are given by
\begin{align}
q &= \mathrm{clip}\!\left(\mathrm{round}\!\left(\frac{w}{\Delta}+z\right),\;0,\;Q_b\right), 
\qquad
\hat{w} = \Delta\,(q-z),
\label{eq:quant}
\end{align}
where \(\Delta\) and \(z\) are the group-wise scale and zero-point, respectively. Layers assigned \(b_\ell=16\) remain in FP16 and are not quantized.

All methods below use the same quantization operator in Eq.~\eqref{eq:quant}; they differ only in how the layer importance scores \(s_\ell\) are computed and how these scores determine the final precision assignment \(\mathbf{b}\).

\subsection{TAQ-IS: Task-aware quantization via information and stability}
\label{sec:taq}

TAQ-IS estimates task-relevant layer importance scores directly from hidden representations, combining information content and activation stability computed on a small task-specific calibration set.
\paragraph{Layer representations.}
Given an input $x$ tokenized to length $T$, let $H_\ell(x)\in\mathbb{R}^{T\times d}$ denote the hidden states output by layer $\ell$, restricted to valid (non-padding) tokens.
TAQ-IS scores each layer using two calibration-time signals: (i) \emph{information}, quantified via matrix entropy, and (ii) \emph{stability}, measured via activation variance.
\paragraph{Information via matrix entropy.}
We collect a reservoir of $r_\ell$ token vectors from $H_\ell(x)$ across $x\sim\mathcal{D}_{\mathrm{calib}}$ and form a matrix $R_\ell\in\mathbb{R}^{r_\ell\times d}$ (rows are sampled token representations).
Let $Z_\ell = R_\ell - \mathbf{1}\mu_\ell^\top$ be the centered matrix with mean $\mu_\ell=\frac{1}{r_\ell}\sum_{i=1}^{r_\ell} R_{\ell,i}$.
Define the (scaled) covariance
\begin{align}
C_\ell = \frac{1}{r_\ell}\, Z_\ell^\top Z_\ell \in \mathbb{R}^{d\times d}.
\end{align}
Let $\{\lambda_{\ell,i}\}_{i=1}^{d}$ be the eigenvalues of $C_\ell$ (nonnegative), and normalize them as
$p_{\ell,i}=\lambda_{\ell,i}/\sum_j \lambda_{\ell,j}$.
The information score is the entropy of this spectrum:
\begin{align}
\mathrm{Info}_\ell
=
-\sum_{i=1}^{d} p_{\ell,i}\,\log\!\left(p_{\ell,i}+\varepsilon\right).
\label{eq:taq_info}
\end{align}
\paragraph{Stability via activation variance.}
Let $h$ denote a scalar activation entry drawn from $H_\ell(x)$, aggregated over all valid tokens, hidden dimensions, and calibration examples.
We estimate the empirical variance
\begin{align}
\mathrm{Var}_\ell = \mathbb{E}[h^2]-\big(\mathbb{E}[h]\big)^2,
\qquad
\mathrm{Stab}_\ell = -\mathrm{Var}_\ell,
\label{eq:taq_stab}
\end{align}
so larger $\mathrm{Stab}_\ell$ corresponds to more stable (lower-variance) activations. We z-normalize both signals across layers, $z(\cdot)$, and combine them:
\begin{align}
s_\ell
=
\alpha\, z(\mathrm{Info}_\ell) + \beta\, z(\mathrm{Stab}_\ell),
\label{eq:taq_score}
\end{align}
with $\alpha,\beta\ge 0$ (we use $\alpha=\beta=0.5$).
We then assign 8-bit to the top $K\%$ layers by $s_\ell$ and 4-bit to the remaining layers. This yields a task-conditioned mixed-precision model without requiring task labels beyond the calibration prompts.
\paragraph{Residual-stream stability under quantization.}
TAQ-IS scores should remain meaningful when layers are quantized jointly, not only in isolated per-layer analysis. To test this, we compute TAQ-IS scores on the clean FP16 model, quantize the first eight transformer blocks to 4-bit, and recompute the downstream scores. As shown in \autoref{fig:motivation_quantization} for Qwen2.5-7B/TriviaQA, the clean and contaminated scores remain closely aligned. This also suggests that TAQ-IS captures a stable task signal in the residual stream and identifies the layers most relevant to the task, rather than relying on unrelated artifacts. We provide a fuller analysis in \autoref{exp:inter_layer_analysis}.

\subsection{TAQ-O: Task-aware quantization via oracle layer sensitivity}
\label{sec:taqo}

TAQ-O is a label-informed diagnostic reference, not a deployable quantization policy. It estimates local layer sensitivity by measuring how much a task metric changes when a single layer is quantized while all other layers remain in FP16
\paragraph{Per-layer sensitivity.}
Let $\mathcal{M}(\cdot)$ be a task metric (e.g., averaged EM/F1) computed on a small held-out subset from $\mathcal{D}_{\mathrm{calib}}$, and let $\mathcal{S}_{\mathrm{base}}=\mathcal{M}(f_\theta)$ be the baseline score in FP16. For each layer $\ell$, define a model $f^{(\ell)}$ where \emph{only} layer $\ell$ is quantized to low precision (4-bit) and all other layers remain FP16; the oracle drop is
\begin{align}
\label{eq:taqo_delta}
f^{(\ell)} = f_{\theta(\,b_\ell=4,\ b_{j\neq \ell}=16\,)}, \qquad
\Delta_\ell
=
\max\!\left(0,\ \mathcal{S}_{\mathrm{base}} - \mathcal{M}\!\left(f^{(\ell)}\right)\right).
\end{align}
Larger $\Delta_\ell$ indicates a layer whose low-bit quantization harms task performance more. In practice, $\Delta_\ell$ is estimated over a fixed held-out calibration subset to ensure consistent comparisons across layers.
\paragraph{Allocation.} We keep a small set of edge layers $\mathcal{E}$ (first and last blocks) in FP16 for robustness, and additionally keep the top-$k$ layers by $\Delta_\ell$ in FP16.
All remaining layers are quantized to 4-bit.
Because TAQ-O evaluates layers marginally and uses task labels, it should not be interpreted as a global optimum or theoretical upper bound under joint quantization. We include it only to compare label-free proxies against a task-metric-aligned layer ranking.


\subsection{TAQ-KL: Task-aware quantization via output-sensitive KL scoring}
\label{sec:taqkl}
TAQ-KL estimates layer importance by how strongly perturbations at a given layer change the model's output distribution, measured via KL divergence.
TAQ-KL provides a label-free alternative to TAQ-O. Rather than evaluating task-metric degradation after quantizing each layer, it injects a small quantization-like perturbation at one layer at a time and measures the induced change in the model's next-token distribution. Thus, TAQ-KL is task-conditioned through the calibration prompts but does not require task labels.
\paragraph{Noise model and logits.}
For an input $x$, let $z(x)\in\mathbb{R}^{|\mathcal{V}|}$ be the baseline logits at the final position.
For each layer $\ell$, we inject additive noise into its hidden states to mimic quantization error:
\noindent
\begin{minipage}{0.47\linewidth}
\begin{equation}
H'_\ell(x) = H_\ell(x) + \eta
\label{eq:taqos_noise_eq1}
\end{equation}
\end{minipage}
\hfill
\begin{minipage}{0.47\linewidth}
\begin{equation}
\eta \sim \mathcal{U}\!\left(-\frac{\delta_\ell}{2},\frac{\delta_\ell}{2}\right)
\label{eq:taqos_noise_eq2}
\end{equation}
\end{minipage}
where $\delta_\ell$ is a layer-dependent scale.
We set $\delta_\ell$ using a simple range-based proxy: let $r_\ell$ be the average per-example range of the last-token hidden state at layer $\ell$ over calibration prompts, then $\delta_\ell \approx r_\ell/(2^4-1)$, approximating the step size of a 4-bit uniform quantizer.
Let $z'_\ell(x)$ denote the logits produced under the forward pass injected with noise.
\paragraph{KL-based sensitivity.}
With temperature $T>0$, define
\[
p(x)=\mathrm{softmax}\!\left(\frac{z(x)}{T}\right), \quad
q_\ell(x)=\mathrm{softmax}\!\left(\frac{z'_\ell(x)}{T}\right).
\]
We score each layer by expected KL divergence over calibration prompts:
\begin{align}
s_\ell
=
\mathbb{E}_{x\sim\mathcal{D}_{\mathrm{calib}}}
\left[
\mathrm{KL}\!\left(p(x)\ \|\ q_\ell(x)\right)
\right].
\label{eq:taqos_kl}
\end{align}
Higher $s_\ell$ indicates that perturbations at layer $\ell$ induce larger changes in the output distribution, suggesting a greater need for higher precision at that layer. As in TAQ-IS, we allocate 8-bit to the top $K\%$ layers by $s_\ell$ and 4-bit to the rest:

%% file: figs/motivation_layers.tex
\begin{figure}[t]
    \centering
    \begin{minipage}{0.50\linewidth}
        \centering
        \includegraphics[width=\linewidth]{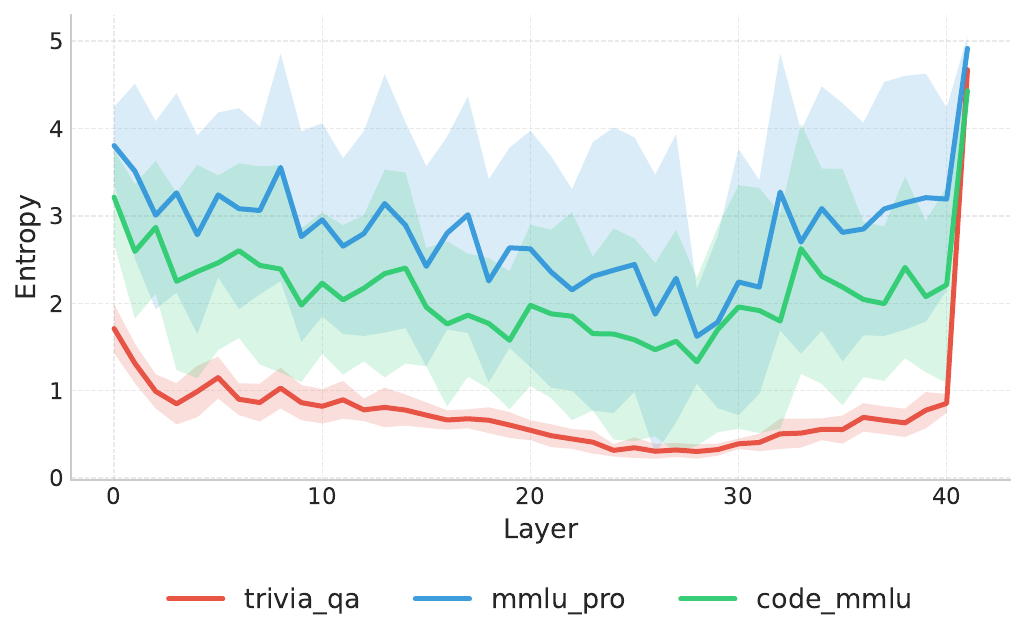}
        \caption{Information entropy across layers with error bars on three tasks(\emph{Gemma2-9B}).}
        \label{fig:motivation_layers}
    \end{minipage}
    \hfill
    \begin{minipage}{0.45\linewidth}
        \centering
        \includegraphics[width=\linewidth]{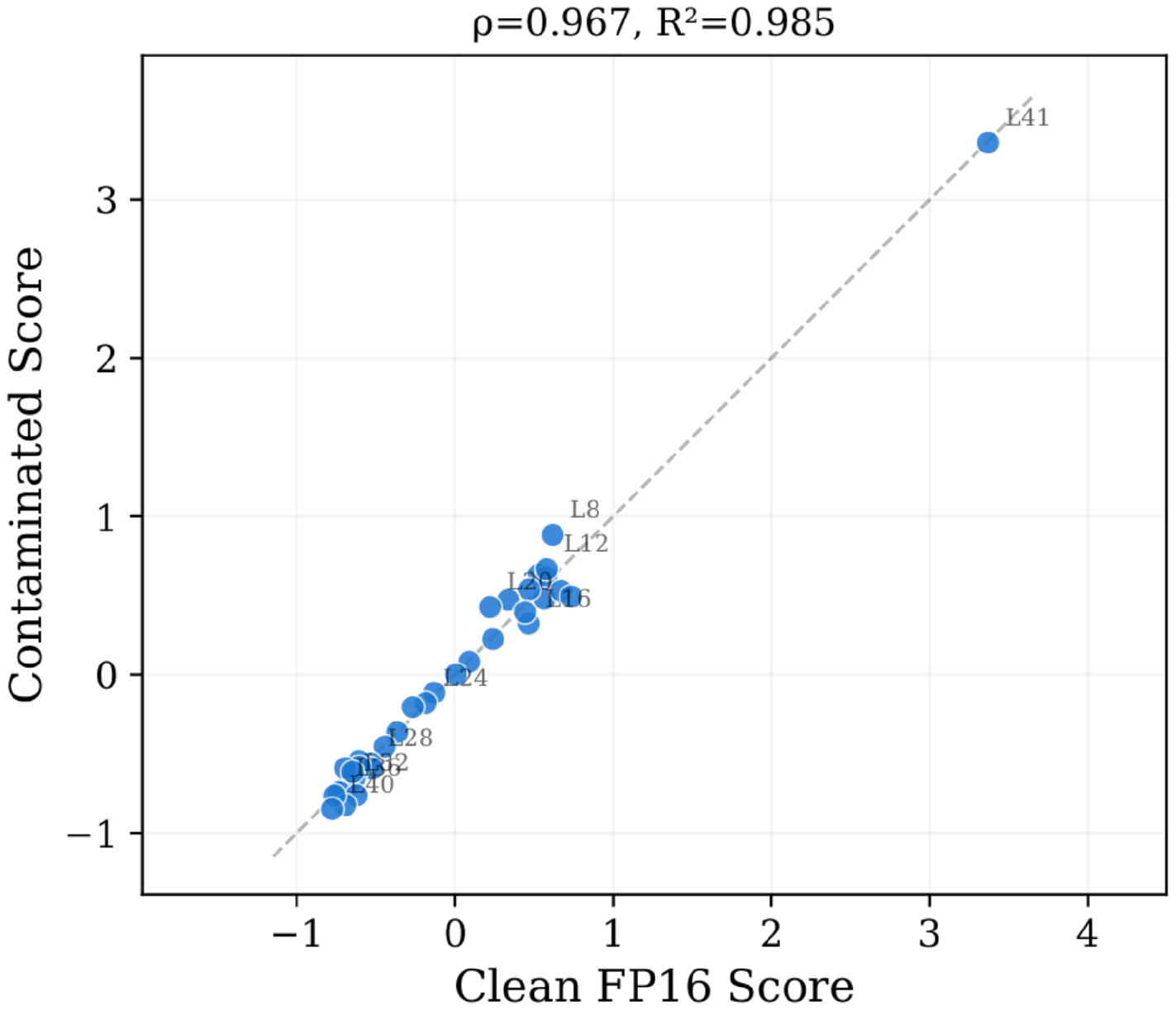}
        \caption{Task-relevant score after (contaminated) quantization.}
        \label{fig:motivation_quantization}
    \end{minipage}
\end{figure}

%% file: sec/5_experiments.tex
\section{Experiments}
\label{sec:experiments}
We evaluate \emph{per-task} PTQ for LLMs through four questions:
\begin{itemize}
    \item \textbf{RQ1.} Does protecting task-relevant layers improve the accuracy--memory trade-off over standard task-agnostic PTQ baselines?
    \item \textbf{RQ2.} Can TAQ remain competitive with, or outperform, a stronger less-constrained mixed-precision baseline whose realized precision budget can be higher?
    \item \textbf{RQ3.} Do label-free signals recover oracle-like layer rankings for precision allocation?
    \item \textbf{RQ4.} Does TAQ remain effective under imperfect or mixed-task calibration?
\end{itemize}

\subsection{Experimental Setting}
\label{sec:settings}

\textbf{Tasks and metrics.} We evaluate performance across three diverse domains spanning knowledge retrieval, code understanding, and mathematical reasoning: (i) TriviaQA (Knowledge Retrieval) \citep{joshi2017triviaqa}, reporting Exact Match (EM) and token-level F1; (ii) CodeMMLU (Code Understanding) \citep{nguyen2025codemmlu}, reporting EM; and (iii) MMLU-Pro (Math/Reasoning) \citep{wang2024mmlu}, reporting EM. \textbf{Models.} We evaluate four open-weight instruction-tuned LLMs spanning two model families and several parameter scales: Qwen2.5-3B-Instruct, Qwen2.5-7B-Instruct \citep{hui2024qwen2}, Qwen2-7B-Instruct \citep{yang2024qwen2technicalreport}, and Gemma-2-9B-it \citep{team2024gemma}. \textbf{Baselines and comparison regimes.} We benchmark against representative PTQ baselines under three complementary regimes. (i) Standard uniform low-bit PTQ baselines, GPTQ \citep{frantar2022gptq} and AWQ \citep{lin2024awq}, which represent canonical 4-bit deployment. (ii) A stronger mixed-precision reference, SliM-LLM \citep{huang2024slim}, which has greater allocation freedom and, in most cases, a higher realized precision budget than most TAQ variants. (iii) Structural controls (Last25, Uniform 8-bit; \autoref{app:additional_baselines}) . We also include W\textbackslash O (FP16) as the full-precision reference. All methods use identical prompts, splits, metric code, and unified realized-weight-footprint accounting (\autoref{app:hardware}); we compare them under realistic accuracy--memory (GB) budgets. \textbf{Quantization protocol.} We focus on weight-only quantization using group-wise affine quantization with a group size of $G{=}128$. Per-layer bitwidths are selected from $\{4, 8, 16\}$, where 16 denotes FP16 retention. The methods differ strictly in their layer ranking and precision allocation policies, as detailed in \S\ref{sec:method}. \textbf{Calibration data.} For each benchmark, we construct disjoint calibration ($D_{\mathrm{calib}}$) and test sets containing 512 and 2{,}048 examples, respectively. $D_{\mathrm{calib}}$ is used to compute layer-importance scores (TAQ-IS, TAQ-KL), derive oracle sensitivities (TAQ-O), and fit quantizer parameters; no gradient-based updates are applied. For TriviaQA, we use a fixed 512-example subset of the validation split for calibration and a disjoint 2{,}048-example subset for evaluation. For MMLU-Pro and CodeMMLU, we use stratified subsets from the official evaluation splits; calibration and evaluation examples are disjoint and sampled with a fixed seed. \textbf{Memory reporting.} We report $W$ as the realized weight footprint in GB under the unified accounting in \autoref{app:hardware}: quantized transformer-linear weights, quantization metadata, FP16-retained blocks, embeddings/\texttt{lm\_head}, and norms. Average bitwidths are reported over transformer-linear layers. \textbf{TAQ implementation details.} \textbf{(1) TAQ-IS} ranks layers using $s_\ell=\alpha z(\mathrm{Info}_\ell)+\beta z(\mathrm{Stab}_\ell)$, where $\mathrm{Stab}_\ell=-\mathrm{Var}_\ell$, with $\alpha=\beta=0.5$, using a reservoir of $r=256$ token vectors per layer. It assigns 8-bit precision to the top $K{=}25\%$ of layers and 4-bit to the remainder. \textbf{(2) TAQ-O} establishes an oracle baseline by quantizing layers individually to 4-bit and retaining FP16 for the top-$k{=}8$ most sensitive layers (plus the first/last 2 layers), using $n_{\mathrm{sens}}=16$ held-out calibration examples. \textbf{(3) TAQ-KL} scores layers via the KL divergence between baseline and perturbed outputs (temperature $T{=}1$) under uniform noise injection $\eta\!\sim\!\mathcal{U}(-\Delta_\ell/2,\Delta_\ell/2)$. It promotes the top $K{=}25\%$ layers to 8-bit. As baselines have different bit allocations and thus a structural advantage, we report Pareto comparisons across the accuracy–memory frontier rather than budget-matched sweeps.

\subsection{Experimental results}
\label{sec:results}

\begin{table*}[t!]
\centering
\caption{\textbf{Main results.} \textbf{W} is the per-model quantized weight size (GB).
\textbf{Bold} marks the best and \underline{underlined} the second-best quantized result.
TriviaQA results are deferred to \autoref{app:trivia_full}.}
\vspace{0.35em}
\label{tab:final_results}

\resizebox{\textwidth}{!}{%
\begin{tabular}{l *{8}{c} | *{8}{c}}
\toprule
\multirow{3}{*}{\textbf{Method}}
& \multicolumn{8}{c|}{\textbf{Math Reasoning: MMLU-Pro (EM$\uparrow$)}}
& \multicolumn{8}{c}{\textbf{Code Understanding: CodeMMLU (EM$\uparrow$)}} \\
\cmidrule(lr){2-9} \cmidrule(lr){10-17}
& \multicolumn{2}{c}{\small Qwen2.5-3B}
& \multicolumn{2}{c}{\small Qwen2.5-7B}
& \multicolumn{2}{c}{\small Qwen2-7B}
& \multicolumn{2}{c|}{\small Gemma-2-9B}
& \multicolumn{2}{c}{\small Qwen2.5-3B}
& \multicolumn{2}{c}{\small Qwen2.5-7B}
& \multicolumn{2}{c}{\small Qwen2-7B}
& \multicolumn{2}{c}{\small Gemma-2-9B} \\
\cmidrule(lr){2-3} \cmidrule(lr){4-5} \cmidrule(lr){6-7} \cmidrule(lr){8-9}
\cmidrule(lr){10-11} \cmidrule(lr){12-13} \cmidrule(lr){14-15} \cmidrule(lr){16-17}
& \small EM$\uparrow$ & \small W$\downarrow$
& \small EM$\uparrow$ & \small W$\downarrow$
& \small EM$\uparrow$ & \small W$\downarrow$
& \small EM$\uparrow$ & \small W$\downarrow$
& \small EM$\uparrow$ & \small W$\downarrow$
& \small EM$\uparrow$ & \small W$\downarrow$
& \small EM$\uparrow$ & \small W$\downarrow$
& \small EM$\uparrow$ & \small W$\downarrow$ \\
\midrule
W\textbackslash O (FP16)
& 37.50 & 5.75 & 33.20 & 14.19 & 41.06 & 14.19 & 42.48 & 17.21
& 50.29 & 5.75 & 48.78 & 14.19 & 51.32 & 14.19 & 53.52 & 17.21 \\
\midrule
GPTQ
& 13.57 & 1.93 & 18.51 & 5.19 & 23.83 & 5.19 & 40.63 & 5.75
& 24.66 & 1.93 & 34.91 & 5.19 & 35.16 & 5.19 & \textbf{51.27} & 5.75 \\
AWQ
& 09.86 & 2.50 & 23.54 & 5.19 & 29.15 & 5.19 & \underline{41.60} & 5.74
& 20.61 & 2.50 & 40.28 & 5.19 & 40.09 & 5.19 & 50.68 & 5.74 \\
SliM-LLM
& 30.86 & 2.77 & 31.01 & 5.83 & 38.48 & 5.83 & 39.89 & 5.83
& 47.04 & 2.77 & 47.17 & 5.83 & \textbf{49.32} & 5.83 & 50.44 & 8.31 \\
\midrule
\textbf{TAQ-IS (Ours)}
& \underline{30.76} & 2.25 & \textbf{33.35} & 5.95 & 37.99 & 5.95 & \underline{41.60} & 6.76
& 46.29 & 2.25 & \underline{47.22} & 5.95 & 48.49 & 5.95 & \underline{51.03} & 6.76 \\
\textbf{TAQ-O (Ours)}
& \textbf{33.01} & 3.20 & \underline{32.13} & 9.05 & \underline{38.57} & 9.05 & 39.40 & 9.02
& \textbf{47.41} & 3.20 & \textbf{49.51} & 9.05 & \underline{49.22} & 9.05 & 50.54 & 9.02 \\
\textbf{TAQ-KL (Ours)}
& 28.03 & 2.25 & \textbf{35.35} & 5.95 & \textbf{39.11} & 5.95 & \textbf{42.48} & 6.76
& \underline{46.83} & 2.25 & 47.02 & 5.95 & 47.12 & 5.95 & 50.83 & 6.76 \\
\bottomrule
\end{tabular}}
\end{table*}

\autoref{tab:final_results} compares TAQ-IS, TAQ-O, and TAQ-KL against FP16, GPTQ, AWQ, and SliM-LLM on MMLU-Pro and CodeMMLU. We additionally evaluate on a knowledge-retrieval benchmark (TriviaQA); the corresponding full results table and exact bit allocations are deferred to \autoref{app:trivia_full} and \autoref{app:layerwise-precision-allocations}, respectively.
\paragraph{Accuracy--memory trade-off vs.\ task-agnostic PTQ (RQ1).} On MMLU-Pro and CodeMMLU, the best TAQ variant improves the accuracy--memory trade-off over standard task-agnostic baselines on every backbone. On CodeMMLU/Qwen2.5-3B, TAQ-KL reaches $46.83\%$ EM at $2.25$\,GB, compared to GPTQ's $24.66\%$ at $1.93$\,GB and AWQ's $20.61\%$ at $2.50$\,GB. On MMLU-Pro/Qwen2.5-7B, TAQ-KL reaches $35.35\%$ EM at $5.95$\,GB, outperforming AWQ by nearly $12$ points with only a $0.76$\,GB premium over AWQ's $5.19$\,GB. Similar patterns hold on Qwen2-7B and Gemma-2-9B. Pareto views of accuracy vs.\ memory are provided in \autoref{subsec:memory_performance_tradeoff}.
\textbf{Comparison against a stronger mixed-precision reference (RQ2).} SliM-LLM is a less-constrained mixed-precision baseline with greater allocation freedom than the fixed TAQ-IS/TAQ-KL policy and, in our setup, a substantially higher realized precision budget. TAQ-IS/TAQ-KL use a constrained $4/8$-bit policy in which only the top $25\%$ of layers receive $8$-bit precision, yielding an average of $5.00$--$5.05$ bits per transformer layer. Under the linear-weight effective-bit accounting, SliM-LLM uses approximately $7.17$, $6.13$, $6.14$, and $7.39$ bits on Qwen2.5-3B, Qwen2.5-7B, Qwen2-7B, and Gemma-2-9B, respectively, while TAQ-IS/TAQ-KL use approximately $4.87$, $4.60$, $4.61$, and $5.05$ bits (\autoref{tab:precision_stats}). SliM-LLM is therefore a deliberately challenging reference: more allocation freedom and a higher effective bit budget. Despite this asymmetry, TAQ remains competitive with or better than SliM-LLM on the structured tasks: on MMLU-Pro, the best TAQ variant outperforms SliM-LLM on every backbone ($33.01$ vs.\ $30.86$ on Qwen2.5-3B; $35.35$ vs.\ $31.01$ on Qwen2.5-7B; $39.11$ vs.\ $38.48$ on Qwen2-7B; $42.48$ vs.\ $39.89$ on Gemma-2-9B); on CodeMMLU, TAQ improves over SliM-LLM on Qwen2.5-3B, Qwen2.5-7B, and Gemma-2-9B, and lands within $0.10$\,EM on Qwen2-7B. Overall, the best TAQ variant matches or exceeds SliM-LLM in $7$ of the $8$ MMLU-Pro/CodeMMLU model--task pairs while spending roughly $1$--$2$ fewer effective bits per linear layer, indicating that the gains do not stem from greater bit budget or allocation freedom but from \emph{where} precision is spent.\\\\
\textbf{Label-free proxies vs.\ oracle (RQ3).} Label-free TAQ-IS and TAQ-KL closely track the label-informed TAQ-O reference. On CodeMMLU/Qwen2.5-3B, TAQ-KL nearly matches TAQ-O while using less memory ($46.83\%$ vs.\ $47.41\%$; $2.25$\,GB vs.\ $3.20$\,GB), and on MMLU-Pro/Qwen2.5-7B TAQ-KL exceeds TAQ-O ($35.35\%$ vs.\ $32.13\%$ at lower memory), consistent with TAQ-O being a marginal-sensitivity diagnostic rather than a true upper bound under joint quantization. Together, these results indicate that representation- and output-sensitivity signals are useful proxies for layer importance without requiring labeled validation data.
\paragraph{Ablation studies.}
\label{sec:sensitivity_ablation}
We test whether TAQ is robust to three practical choices: calibration size, bit budget, and scoring signal. First, we vary the calibration size, $|D_{\mathrm{calib}}| \in \{64,128,256,512,1024\}$, and find that EM changes by only $\sim$1.3 points on Qwen2.5-7B/MMLU-Pro, showing that small calibration sets are sufficient (\autoref{tab:appx_calib_size}). Second, we vary the bit budget by promoting the top $K\% \in \{10,25,50\}\%$ layers to 8-bit. Performance stays in a narrow range: $K{=}25\%$ gives the best MMLU-Pro result, while $K{=}10\%$ is slightly better on CodeMMLU. Since increasing the budget to $50\%$ does not improve performance, the benefit is not explained by simply protecting more layers; selecting the right layers matters, so we use $K{=}25\%$ as a strong default trade-off (\autoref{tab:appx_k_sweep}). Some TAQ configurations exceed FP16 accuracy; we interpret these cases cautiously. A possible explanation is that task-conditioned allocation preserves precision in layers most relevant to the calibration task while compressing less relevant layers more aggressively, acting as an implicit regularizer. This may also explain why performance can drop when $K$ is too high.  Finally, we test which scoring signals best select the 8-bit layers. For TAQ-IS, we compare entropy-only, variance-only, and the default combined score with the model, calibration size, and $K{=}25\%$ fixed. Variance-only works best on MMLU-Pro, entropy-only works best on CodeMMLU, and the combined score is most stable overall. Across TAQ scoring families, different tasks favor different signals, with TAQ-KL leading on reasoning and TAQ-O leading on code (\autoref{tab:appx_alpha_sweep}, \autoref{tab:appx_metric_ablation}).
\paragraph{Additional baselines and controls.}
\label{sec:additional_baselines}
We add two control baselines to separate the effect of task-aware layer selection from the effect of simply using more precision. First, \textbf{Last25} uses the same $4/8$-bit budget as TAQ-IS and TAQ-KL, but assigns 8-bit precision to the last $25\%$ of layers. Second, \textbf{Uniform 8-bit} assigns 8-bit precision to every layer, providing a higher-memory reference for the effect of increasing precision globally. The results show that task-aware selection matters. On MMLU-Pro, at least one TAQ variant outperforms Last25 on every backbone, for example $35.35$ vs.\ $31.35$ on Qwen2.5-7B and $42.48$ vs.\ $40.97$ on Gemma-2-9B. Compared with Uniform 8-bit, TAQ is not always higher in raw EM, but it is often competitive while using substantially less memory, and on Qwen2.5-7B and Gemma-2-9B the best TAQ variant also exceeds the available Uniform 8-bit result. Together, these controls show that TAQ's improvements are not explained simply by assigning 8-bit precision to the last layers or by using uniformly higher precision; the task-aware choice of protected layers is important. Full results are reported in \autoref{tab:additional_baselines}.
\paragraph{Mixed-task calibration.}
\label{sec:robustness_tasks}
We next test whether TAQ remains effective on mixed-task calibration data, addressing \textbf{RQ4}. We consider two settings. First, instead of computing a separate TAQ-IS allocation for each benchmark, we compute one allocation on Qwen2.5-7B using a mixed calibration set of $512$ prompts sampled from TriviaQA, CodeMMLU, and MMLU-Pro. We then evaluate this same allocation separately on each benchmark. The mixed allocation improves over the single-task allocation in our run: CodeMMLU increases from $47.22$ to $49.61$ EM, MMLU-Pro from $33.35$ to $34.28$ EM, and TriviaQA from $12.55$ to $17.09$ EM (\autoref{tab:mixed_task}). We hypothesize that using a broader calibration pool can reduce overfitting.
\paragraph{Cross-task allocation stability.} Second, we check how much the selected layers actually change across tasks and models. Across the $5{\times}3$ model/dataset grid, $87.5\%$ of layer assignments remain the same across calibration tasks, with cross-task Spearman correlation $\rho=0.757$. The layers that do change are mostly near the first and last transformer blocks, consistent with the boundary-layer pattern in \autoref{fig:appx_mixed_heatmap}. Together, these results suggest that many important layers are shared across tasks, while task-specific differences are concentrated in a small number of layers. Thus, TAQ can often produce a robust shared allocation.
\paragraph{Calibration source vs.\ allocation policy.}
\label{sec:calib_vs_policy}
We also test whether the gains come from task-aligned calibration data alone, or from the task-aware allocation policy itself. In \autoref{app:calib_vs_policy}, we evaluate on a broader AWQ-supported model suite: Phi-4, Qwen3, Llama-3.1, Qwen2.5, and Mistral. We compare three settings: \textbf{AWQ-Mixed}, where AWQ is calibrated on a mixed pool from three tasks; \textbf{AWQ-OT}, where AWQ is calibrated only on the target task; and our \textbf{TAQ-IS} and \textbf{TAQ-O} task-aware allocation policies. If task-aligned calibration alone were sufficient, AWQ-OT should consistently improve over AWQ-Mixed. Instead, AWQ-OT is unstable: it improves Llama-3.1, but degrades Phi-4 and Qwen2.5, and changes Mistral only slightly. In contrast, the best quantized results come from TAQ-IS or TAQ-O.
\paragraph{Hardware and deployment validation.}
\label{sec:hardware_validation}
We validate TAQ with real quantized kernels on a single NVIDIA A40, comparing against FP16/FP32, uniform 4-bit, GPTQ-Int4, and AWQ-Int4 across three backbones and two batch sizes. On Qwen2.5--7B at batch size $1$, TAQ-IS improves throughput from $27.63$ to $37.29$ tok/s relative to FP16, giving $+35\%$ throughput and $-26\%$ latency; the gain remains positive at batch size $8$ ($+11.6\%$; \autoref{app:hardware}). TAQ-IS also outperforms uniform 4-bit ($28.71$ tok/s), GPTQ-Int4 ($4.40$ tok/s), and AWQ-Int4 ($16.12$ tok/s). Under the unified memory accounting in \autoref{app:hardware}, TAQ-IS/TAQ-KL use $5.95$\,GB on Qwen2.5--7B, a $0.76$\,GB premium over uniform 4-bit GPTQ/AWQ ($5.19$\,GB), while preserving 8-bit precision on task-sensitive layers. With real NF4/Int8 kernels, Qwen2.5--3B realizes $2.20$\,GB weight size and $2.31$\,GB peak memory, corresponding to $2.6\times$ compression over FP16. The slight difference from the accounting-based $W$ reported in \autoref{tab:final_results} ($2.25$\,GB for Qwen2.5--3B TAQ-IS/TAQ-KL) comes from backend-specific packing in the realized NF4/Int8 kernels. Full throughput, latency, and memory breakdowns are in \autoref{app:hardware}.\\\\
\textbf{Inter-layer score stability and residual-stream propagation.}\label{exp:inter_layer_analysis}
We test whether layer scores from clean FP16 activations remain reliable after upstream 4-bit quantization perturbs the residual stream. Since TAQ uses these scores mainly to choose higher-precision layers, we evaluate rank preservation via Spearman's $\rho$ between clean FP16 rankings and rankings recomputed after quantizing the first eight transformer blocks. Rankings remain highly stable on TriviaQA, with combined-score correlations of $\rho{=}0.997/0.999/0.990$ on Qwen2.5--7B and $\rho{=}0.967/0.969/0.952$ on Gemma-2--9B (TAQ-IS/TAQ-O/TAQ-KL); Qwen2.5--7B is also stable on CodeMMLU ($\rho{\approx}0.75$--$0.80$). The main weaker case is Gemma-2--9B on CodeMMLU and MMLU-Pro ($\rho{\approx}0.38$--$0.59$), indicating stronger task-dependent inter-layer interactions where early-layer quantization can alter later-layer rankings. Complementary cosine similarity is also positive: after an initial drop at the cutoff, it stabilizes around $0.93$, so residual-stream perturbations do not grow with depth. Overall, clean FP16 scores are useful proxies for selecting task-sensitive layers (especially on TriviaQA and Qwen2.5--7B). Full results are in \autoref{app:interlayer}.

%% file: sec/7_conclusions.tex
\section{Discussion}
\label{discussion}
We present TAQ, a per-task PTQ framework that uses internal representations to identify task-critical layers and allocate bits accordingly, turning quantization into a \emph{policy} problem rather than a uniform compression step. Across most models and tasks, TAQ yields strong accuracy--memory trade-offs, with especially large gains on specialized workloads where strong baselines can degrade sharply. In some cases, quantized TAQ models even outperform the FP16 model. We view this as a hypothesis that task-aware quantization can preserve task-relevant signal while suppressing irrelevant activation or weight noise, acting as a form of task-conditioned denoising rather than only compression. The label-free variants, TAQ-KL and TAQ-IS, often closely match, and occasionally exceed, the TAQ-O oracle reference, suggesting that intrinsic stability and output-sensitivity signals can recover useful task-aware layer rankings. TAQ-O is therefore best understood as a label-informed diagnostic reference under our evaluation protocol, not as a deployable method or theoretical upper bound. 
\paragraph{Limitations.} TAQ relies on calibration prompts that represent the target deployment distribution; performance may degrade under distribution shift. In addition, TAQ-IS, TAQ-KL, and TAQ-O use tractable local proxies rather than solving the full joint mixed-precision allocation problem, so higher-order interactions among quantized layers may not be fully captured. 
\paragraph{Broader impacts.} TAQ may have positive societal impacts by making LLM inference more efficient and accessible, reducing memory footprint, latency, and potentially energy costs for task-specific deployments. However, more efficient deployment can also lower barriers to using capable LLMs in harmful or sensitive applications, and task-conditioned quantization may degrade behavior under distribution shift or poorly specified calibration data; therefore, TAQ should be evaluated on target-domain safety, privacy, fairness, and robustness criteria before deployment, especially in high-stakes settings. 
\paragraph{Future work.}
Future work should extend TAQ beyond layer-level allocation and model quantization error more directly. Our linearity analysis suggests that quantization error can appear as a persistent residual-stream direction, motivating interpretability-based corrections such as activation steering or task-direction estimation.

\clearpage

%% file: sec/8_appendix.tex
\clearpage
\appendix

\section{Assets and compute}

\subsection{Existing assets and licenses}
\label{subsec:assets_licence}
Our experiments use only public third-party datasets, model checkpoints, and baseline methods, and we do not redistribute third-party data or model weights. The datasets are TriviaQA v1.0 validation data (\url{https://nlp.cs.washington.edu/triviaqa/}; Apache-2.0 code license, with original question/document copyrights retained by their owners), CodeMMLU (\url{https://huggingface.co/datasets/Fsoft-AIC/CodeMMLU}; MIT), and MMLU-Pro (\url{https://huggingface.co/datasets/TIGER-Lab/MMLU-Pro}; MIT dataset license; Apache-2.0 evaluation code). The model checkpoints are Qwen2.5-3B-Instruct (\url{https://huggingface.co/Qwen/Qwen2.5-3B-Instruct}; Qwen Research License), Qwen2.5-7B-Instruct (\url{https://huggingface.co/Qwen/Qwen2.5-7B-Instruct}; Apache-2.0), Qwen2-7B-Instruct (\url{https://huggingface.co/Qwen/Qwen2-7B-Instruct}; Apache-2.0), Gemma-2-9B-it (\url{https://huggingface.co/google/gemma-2-9b-it}; Gemma license / Google usage terms), Phi-4 (\url{https://huggingface.co/microsoft/phi-4}; MIT), Llama-3.1-8B-Instruct (\url{https://huggingface.co/meta-llama/Llama-3.1-8B-Instruct}; Llama 3.1 Community License), and Mistral-7B-Instruct-v0.3 (\url{https://huggingface.co/mistralai/Mistral-7B-Instruct-v0.3}; Apache-2.0). The auxiliary component-analysis suite in \autoref{app:calib_vs_policy} also includes a Qwen3 backbone; the submitted experiment table reports this backbone only at the family level, so the exact Qwen3 checkpoint identifier and license should be inserted from the experiment manifest before final submission rather than inferred from the family name. Baselines are credited to GPTQ, AWQ, and SliM-LLM; we use released code/assets when licensed for research use or our own reimplementation of the published method. All assets are cited to their original creators and used only for research and evaluation in accordance with their licenses and terms.

\subsection{Compute resources}
\label{subsec:compute_resources}
All reported calibration, quantization, evaluation, ablation, robustness, and hardware-validation experiments were run on single-GPU NVIDIA A40 workers with 48 GB of GPU memory. Each model--task--method configuration used one A40 worker and no distributed training, since TAQ is training-free and performs calibration-time scoring, weight-only quantization, and evaluation rather than gradient-based training. A typical configuration required approximately $1.5$--$3$ A40 GPU-hours for 3B models and $3$--$6$ A40 GPU-hours for 7B/9B models, including calibration/scoring, quantization, and evaluation on $2{,}048$ examples. Hardware throughput and latency measurements required approximately $0.5$ A40 GPU-hours per configuration. Across all reported experiments, including baselines, ablations, robustness analyses, and hardware validation, the total compute was approximately $450$ A40 GPU-hours. Including preliminary, exploratory, and failed runs not reported in the paper, the full research project required approximately $750$ A40 GPU-hours. Persistent storage per run was below $100$~GB.

\section{Detailed results}

\subsection{Memory--performance trade-off}
\label{subsec:memory_performance_tradeoff}

\begin{figure}[H]
    \centering
    \includegraphics[page=1,width=\textwidth]{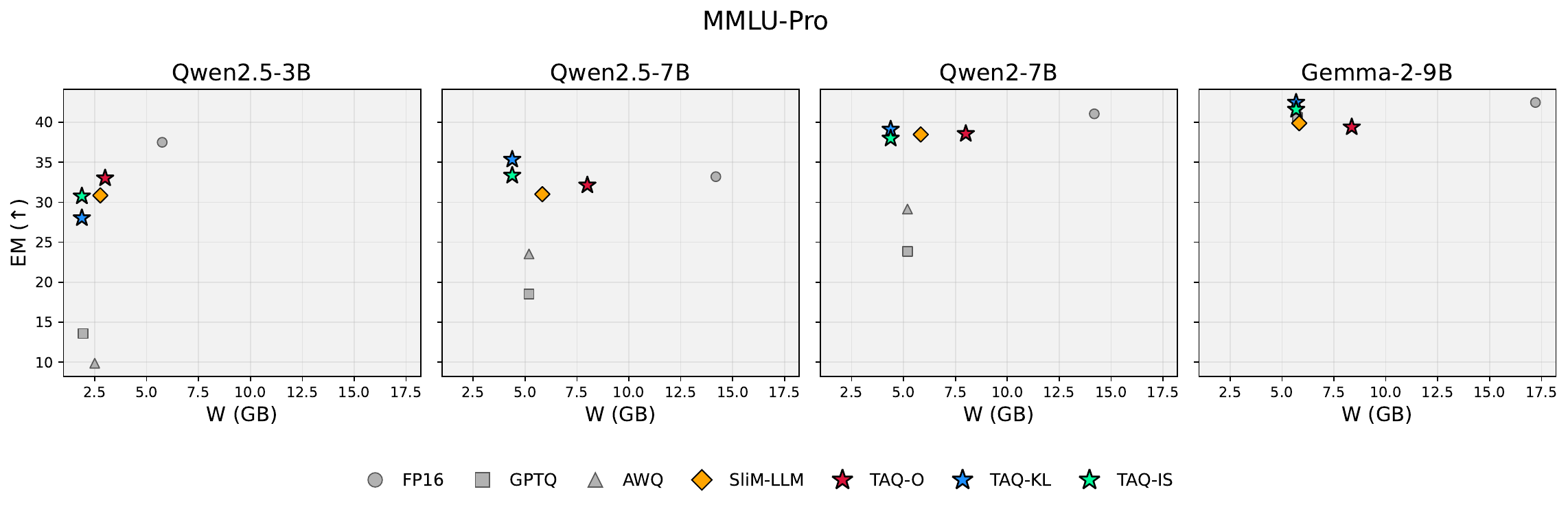}
    \vspace{0.5em}
    \includegraphics[page=2,width=\textwidth]{imgs/all_weight_vs_em_results-2.pdf}
    \vspace{0.5em}
    \includegraphics[page=3,width=\textwidth]{imgs/all_weight_vs_em_results-2.pdf}
    \caption{\textbf{Accuracy--memory trade-off across benchmarks and models.} Each panel corresponds to one benchmark dataset (MMLU-Pro, CodeMMLU, TriviaQA). Within each benchmark, we plot exact match accuracy (EM, $\uparrow$) against the realized weight footprint $W$ in GB for Qwen2.5-3B, Qwen2.5-7B, Qwen2-7B, and Gemma-2-9B. Markers denote FP16, GPTQ, AWQ, SliM-LLM, TAQ-O, TAQ-KL, and TAQ-IS. Points closer to the upper-left indicate better accuracy--memory efficiency.}
    \label{fig:weight_vs_em}
\end{figure}

\subsection{Layerwise precision allocation}
\label{app:layerwise-precision-allocations}

For TAQ-IS and TAQ-KL, the top $25\%$ of layers are assigned 8-bit precision while the remaining layers are assigned 4-bit precision. TAQ-O uses a different diagnostic allocation: the selected sensitive layers are retained in FP16 and the remaining layers are quantized to 4-bit. For visual comparability, the allocation plots show TAQ-O retained layers on the high-precision row; semantically these layers are FP16-retained and are accounted as 16-bit in the memory tables.

\begin{figure}[H]
\centering
\textbf{Qwen2-7B-Instruct}\par
\includegraphics[width=0.95\linewidth]{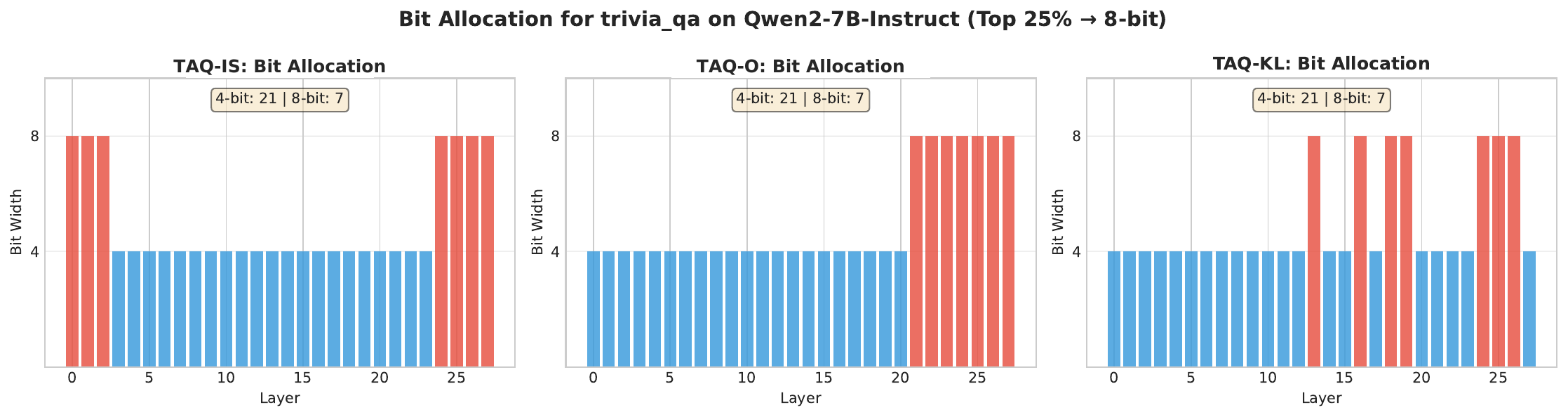}\par
\textbf{gemma-2-9b-it}\par
\includegraphics[width=0.95\linewidth]{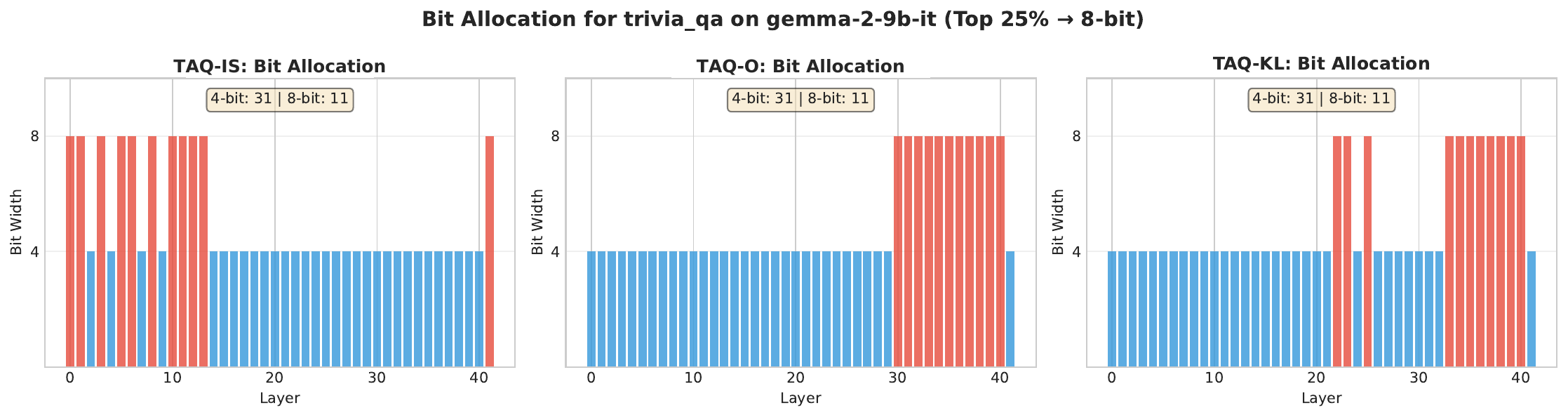}
\caption{Layerwise high-precision allocation on TriviaQA.}
\label{fig:alloc-triviaqa}
\end{figure}

\begin{figure}[H]
\centering
\textbf{Qwen2-7B-Instruct}\par
\includegraphics[width=0.95\linewidth]{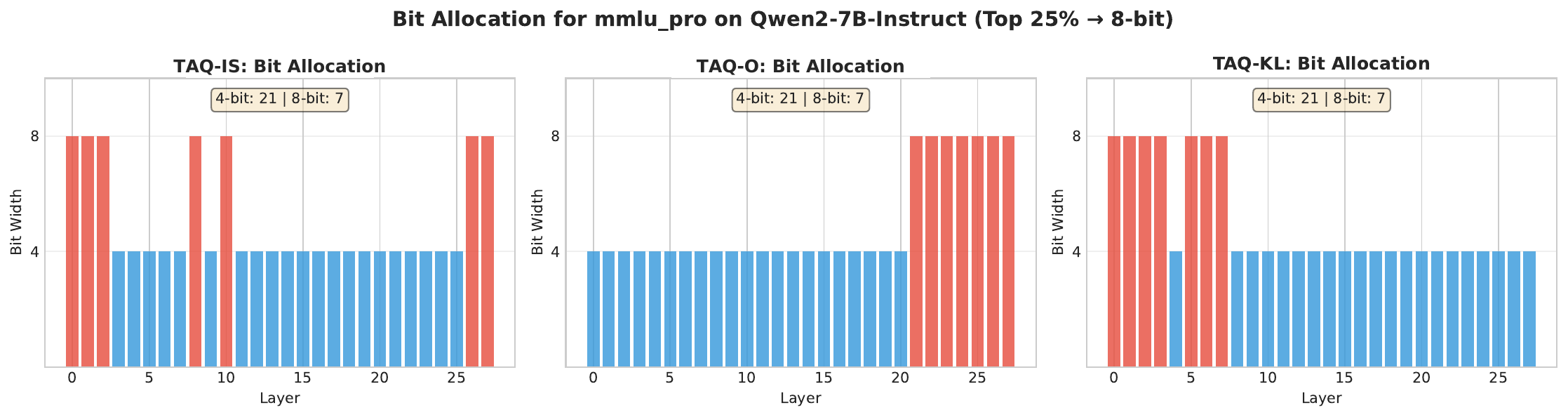}\par
\textbf{gemma-2-9b-it}\par
\includegraphics[width=0.95\linewidth]{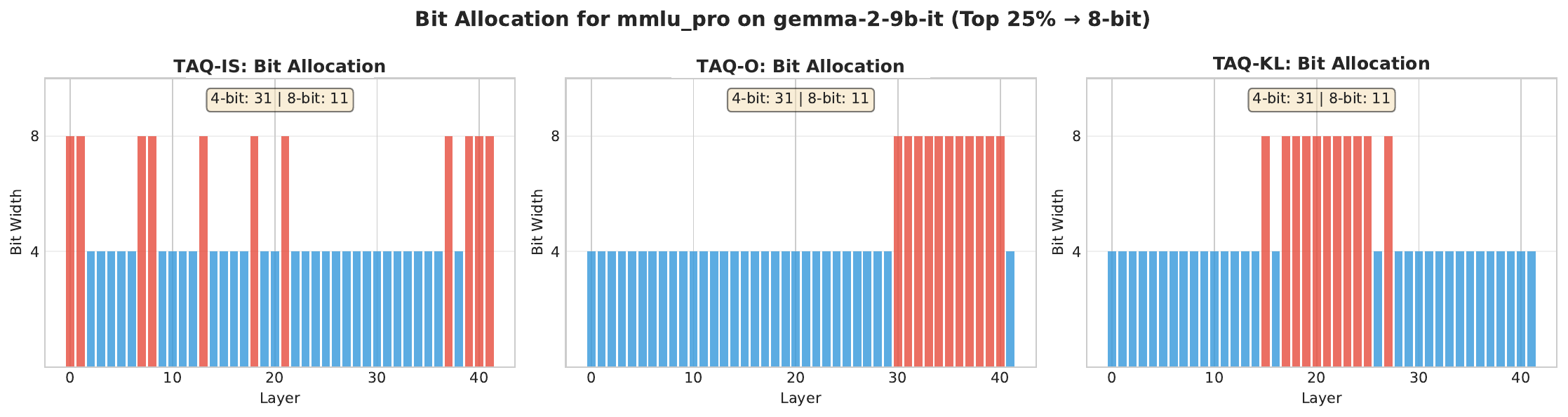}
\caption{Layerwise high-precision allocation on MMLU-Pro.}
\label{fig:alloc-mmlupro}
\end{figure}

\begin{figure}[H]
\centering
\textbf{Qwen2-7B-Instruct}\par
\includegraphics[width=0.95\linewidth]{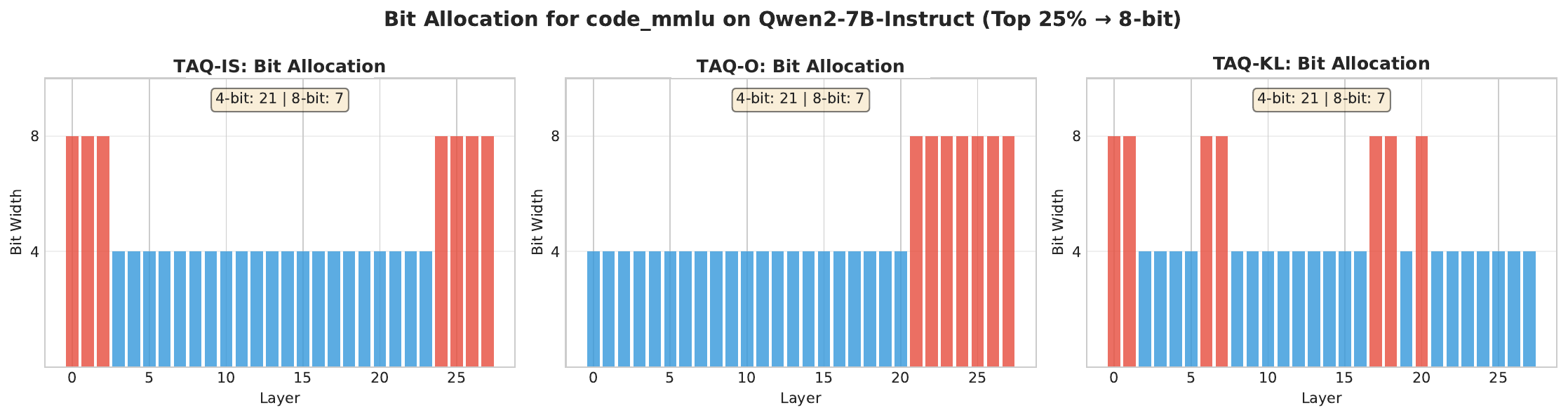}\par
\textbf{gemma-2-9b-it}\par
\includegraphics[width=0.95\linewidth]{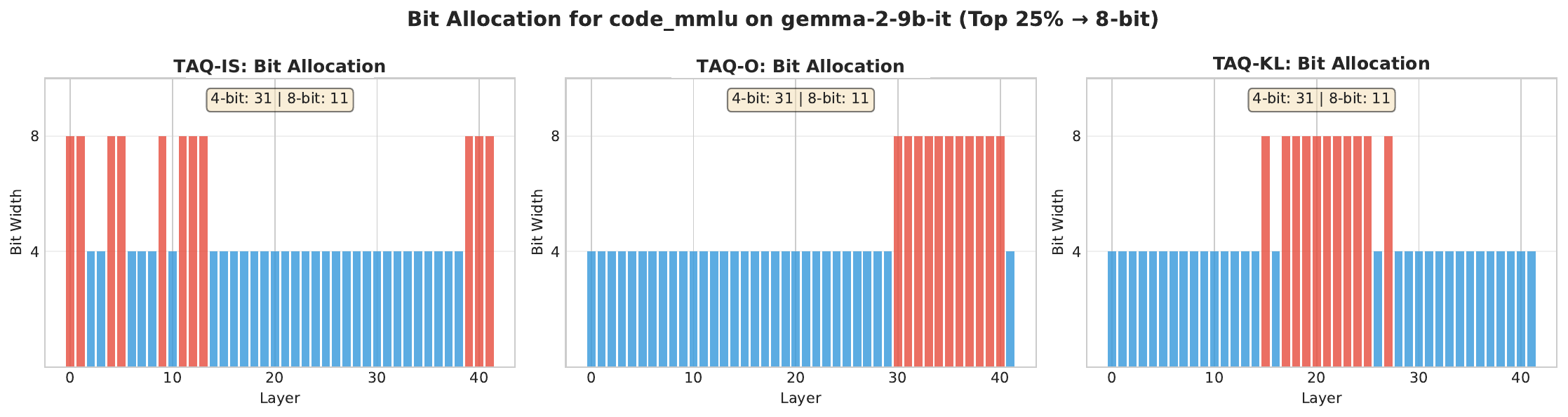}
\caption{Layerwise high-precision allocation on CodeMMLU.}
\label{fig:alloc-codemmlu}
\end{figure}

\subsection{Precision allocation statistics}
\label{app:precision_stats_subsec}

\autoref{tab:precision_stats} reports the average bits per transformer-linear layer and the realized weight footprint (GB) under the unified accounting in \autoref{app:hardware}. Effective-bit numbers are computed on linear weights only (excluding embeddings, \texttt{lm\_head}, and norms). TAQ-IS/TAQ-KL use a constrained $4/8$-bit policy, while SliM-LLM is a less-constrained mixed-precision baseline whose realized precision budget is substantially higher. This contextualizes the main-paper comparison: the best TAQ variant outperforms SliM-LLM on $7$ of $8$ MMLU-Pro/CodeMMLU model--task pairs, and is within $0.10$ EM on the remaining CodeMMLU/Qwen2-7B pair, while spending roughly $1$--$2$ fewer effective bits per linear layer.

\begin{table}[H]
\centering
\caption{\textbf{Precision allocation statistics.} Effective bits per linear layer and realized quantized weight size (GB). TAQ-IS/TAQ-KL use a constrained $4/8$-bit policy; SliM-LLM uses a less-constrained mixed-precision allocation with a higher effective bit budget; TAQ-O retains FP16 on selected layers.}
\label{tab:precision_stats}
\small
\setlength{\tabcolsep}{4pt}
\begin{tabular}{lccccc}
\toprule
\textbf{Method} & \textbf{Qwen2.5-3B} & \textbf{Qwen2.5-7B} & \textbf{Qwen2-7B} & \textbf{Gemma-2-9B} & \textbf{Llama-3.1-8B} \\
\midrule
\multicolumn{6}{l}{\textit{Effective bits per linear layer}} \\
SliM-LLM & 7.17 & 6.13 & 6.14 & 7.39 & -- \\
TAQ-IS   & 4.87 & 4.60 & 4.61 & 5.05 & 5.00 \\
TAQ-KL   & 4.87 & 4.60 & 4.61 & 5.05 & 5.00 \\
TAQ-O    & 7.77 & 8.41 & 8.42 & 7.43 & 8.50 \\
\midrule
\multicolumn{6}{l}{\textit{Realized weight footprint (GB)}} \\
SliM-LLM & 2.77 & 5.83 & 5.83 & 8.31 & -- \\
TAQ-IS   & 2.25 & 5.95 & 5.95 & 6.76 & 5.68 \\
TAQ-KL   & 2.25 & 5.95 & 5.95 & 6.76 & 5.68 \\
TAQ-O    & 3.20 & 9.05 & 9.05 & 9.02 & 8.36 \\
\bottomrule
\end{tabular}
\end{table}

\subsection{Knowledge retrieval: TriviaQA full results}
\label{app:trivia_full}

\autoref{tab:trivia_appx} reports the full TriviaQA results across the four main backbones.
\begin{table}[H]
\centering
\caption{\textbf{Knowledge Retrieval --- TriviaQA full results} (EM$\uparrow$, F1$\uparrow$; $W$ is the realized weight footprint in GB). \textbf{Bold} marks the best and \underline{underlined} the second-best quantized result per metric/backbone. Extended TriviaQA evaluation across additional backbones (Phi-4, Qwen3, Llama-3.1, Mistral) and additional baseline regimes (mixed-calibration AWQ, on-task AWQ-OT) is reported in \autoref{app:calib_vs_policy}.}
\label{tab:trivia_appx}
\resizebox{\textwidth}{!}{%
\begin{tabular}{l *{12}{c}}
\toprule
\multirow{2}{*}{\textbf{Method}}
& \multicolumn{3}{c}{\small Qwen2.5-3B}
& \multicolumn{3}{c}{\small Qwen2.5-7B}
& \multicolumn{3}{c}{\small Qwen2-7B}
& \multicolumn{3}{c}{\small Gemma-2-9B} \\
\cmidrule(lr){2-4} \cmidrule(lr){5-7} \cmidrule(lr){8-10} \cmidrule(lr){11-13}
& EM$\uparrow$ & F1$\uparrow$ & W$\downarrow$
& EM$\uparrow$ & F1$\uparrow$ & W$\downarrow$
& EM$\uparrow$ & F1$\uparrow$ & W$\downarrow$
& EM$\uparrow$ & F1$\uparrow$ & W$\downarrow$ \\
\midrule
W\textbackslash O (FP16) & 17.24 & 24.69 & 5.75 & 16.94 & 25.62 & 14.19 & 00.73 & 12.94 & 14.19 & 62.11 & 68.92 & 17.21 \\
\midrule
GPTQ      & 07.23 & 15.54 & 1.93 & \textbf{15.97} & \textbf{26.57} & 5.19 & \underline{05.71} & 17.10 & 5.19 & 52.64 & 61.04 & 5.75 \\
AWQ       & \textbf{09.72} & \underline{17.66} & 2.50 & 06.98 & 18.15 & 5.19 & \textbf{14.79} & \textbf{25.16} & 5.19 & 51.51 & 60.14 & 5.74 \\
SliM-LLM  & 03.22 & 17.48 & 2.77 & 15.77 & 26.20 & 5.83 & 02.44 & 21.09 & 5.83 & \textbf{61.23} & \textbf{68.87} & 8.31 \\
\midrule
\textbf{TAQ-IS (Ours)} & 03.91 & 17.14 & 2.25 & 12.55 & \underline{23.95} & 5.95 & 02.25 & 20.79 & 5.95 & 59.67 & 68.08 & 6.76 \\
\textbf{TAQ-O  (Ours)} & \underline{08.45} & \textbf{20.34} & 3.20 & 12.26 & 23.23 & 9.05 & 01.86 & 20.58 & 9.05 & 59.33 & 67.67 & 9.02 \\
\textbf{TAQ-KL (Ours)} & 01.32 & 13.69 & 2.25 & \underline{14.06} & 23.83 & 5.95 & 02.25 & \underline{21.17} & 5.95 & \underline{61.04} & \underline{68.58} & 6.76 \\
\bottomrule
\end{tabular}}
\end{table}

\section{Sensitivity and ablation studies}
\label{app:sensitivity_ablation}

This section reports the auxiliary sweeps referenced from \autoref{sec:sensitivity_ablation}. \autoref{tab:appx_metric_ablation} compares the three TAQ scoring families head-to-head on Qwen2.5-7B; \autoref{tab:appx_k_sweep} reports the top-$K\%$ promotion sweep; \autoref{tab:appx_alpha_sweep} reports the $\alpha/\beta$ sweep at $K{=}25\%$; and \autoref{tab:appx_calib_size} reports the calibration-size sensitivity.

\begin{table}[H]
\centering
\caption{\textbf{Per-metric scoring head-to-head} on Qwen2.5-7B (EM$\uparrow$). No single rule dominates across tasks.}
\label{tab:appx_metric_ablation}
\small
\setlength{\tabcolsep}{8pt}
\begin{tabular}{l cc}
\toprule
\textbf{Metric} & \textbf{MMLU-Pro} & \textbf{CodeMMLU} \\
\midrule
TAQ-IS & 33.35 & 47.22 \\
TAQ-O  & 32.13 & \textbf{49.51} \\
TAQ-KL & \textbf{35.35} & 47.02 \\
\bottomrule
\end{tabular}
\end{table}

\begin{table}[H]
\centering
\caption{\textbf{Top-$K\%$ promotion sweep} for TAQ-IS on Qwen2.5-7B (calib$=$512, $G{=}128$). Performance varies within a narrow range; $K{=}25\%$ gives the best MMLU-Pro result and is within one EM point of the best CodeMMLU result in this sweep.}
\label{tab:appx_k_sweep}
\small
\setlength{\tabcolsep}{8pt}
\begin{tabular}{c c cc}
\toprule
\textbf{$K\%$} & \textbf{8-bit layers} & \textbf{MMLU-Pro EM$\uparrow$} & \textbf{CodeMMLU EM$\uparrow$} \\
\midrule
10\% & 3  & 33.25 & \textbf{48.00} \\
25\% & 7  & \textbf{34.62} & 47.22 \\
50\% & 14 & 32.52 & 46.88 \\
\bottomrule
\end{tabular}
\end{table}

\begin{table}[H]
\centering
\caption{\textbf{$\alpha/\beta$ sweep for the TAQ-IS combined score} ($K{=}25\%$ fixed).}
\label{tab:appx_alpha_sweep}
\small
\setlength{\tabcolsep}{8pt}
\begin{tabular}{cc l cc}
\toprule
\textbf{$\alpha$} & \textbf{$\beta$} & \textbf{Signal} & \textbf{MMLU-Pro EM$\uparrow$} & \textbf{CodeMMLU EM$\uparrow$} \\
\midrule
0.0 & 1.0 & Variance only      & \textbf{35.11} & 48.54 \\
0.5 & 0.5 & Combined (default) & 34.62 & 47.22 \\
1.0 & 0.0 & Entropy only       & 30.32 & \textbf{50.44} \\
\bottomrule
\end{tabular}
\end{table}

\begin{table}[H]
\centering
\caption{\textbf{Calibration-size sensitivity} for TAQ-IS on Qwen2.5-7B/MMLU-Pro (eval$=$2{,}048). EM moves within $\sim$1.3 EM points across a $16\times$ range.}
\label{tab:appx_calib_size}
\small
\setlength{\tabcolsep}{8pt}
\begin{tabular}{l ccccc}
\toprule
\textbf{$|D_{\mathrm{calib}}|$} & \textbf{64} & \textbf{128} & \textbf{256} & \textbf{512 (default)} & \textbf{1024} \\
\midrule
TAQ-IS EM$\uparrow$ & 34.18 & \textbf{34.62} & 33.79 & 33.35 & 33.89 \\
\bottomrule
\end{tabular}
\end{table}

\section{Additional baselines and robustness}
\label{app:robustness_appendix}

\subsection{Structural baselines: Last25 and Uniform 8-bit}
\label{app:additional_baselines}

\autoref{tab:additional_baselines} reports per-model EM on MMLU-Pro for the Last25 and Uniform 8-bit structural references introduced in \autoref{sec:additional_baselines}.

\begin{table}[H]
\centering
\caption{\textbf{Additional reference baselines on MMLU-Pro (EM$\uparrow$).} Last25 matches the bit-allocation pattern of TAQ-IS/TAQ-KL by promoting the last $25\%$ of layers; Uniform 8-bit is a higher-memory reference in which all transformer layers receive 8-bit precision. TAQ rows are reproduced from \autoref{tab:final_results} for direct comparison.}
\label{tab:additional_baselines}
\small
\setlength{\tabcolsep}{6pt}
\begin{tabular}{l cccc}
\toprule
\textbf{Method} & \textbf{Qwen2.5-3B} & \textbf{Qwen2.5-7B} & \textbf{Qwen2-7B} & \textbf{Gemma-2-9B} \\
\midrule
Last25 (4/8, last-25\%) & 25.10 & 31.35 & 38.53 & 40.97 \\
Uniform 8-bit           & 37.11 & 33.06 & 40.72 & 42.12    \\
\midrule
TAQ-O  (Ours) & \textbf{33.01} & 32.13 & 38.57 & 39.40 \\
TAQ-KL (Ours) & 28.03 & \textbf{35.35} & \textbf{39.11} & \textbf{42.48} \\
TAQ-IS (Ours) & 30.76 & 33.35 & 37.99 & 41.60 \\
\bottomrule
\end{tabular}
\end{table}

\subsection{Mixed-task calibration and cross-task allocation stability}
\label{app:robustness_tasks}

\autoref{tab:mixed_task} reports the single-task vs.\ mixed-task EM comparison on Qwen2.5-7B; mixed-task calibration improves over single-task on every benchmark in this run. \autoref{fig:appx_mixed_heatmap} is the bit-allocation heatmap aggregated across the $5{\times}3$ model/dataset grid.

\begin{table}[H]
\centering
\caption{\textbf{Mixed-task calibration} on Qwen2.5-7B (TAQ-IS, calib$=$512, eval$=$2{,}048). A single allocation derived from a uniform pool of all three tasks is evaluated per benchmark.}
\label{tab:mixed_task}
\small
\setlength{\tabcolsep}{8pt}
\begin{tabular}{l ccc}
\toprule
\textbf{Eval Dataset} & \textbf{Single-task EM} & \textbf{Mixed-task EM} & \textbf{$\Delta$} \\
\midrule
CodeMMLU & 47.22 & \textbf{49.61} & $+2.39$ \\
MMLU-Pro & 33.35 & \textbf{34.28} & $+0.93$ \\
TriviaQA & 12.55 & \textbf{17.09} & $+4.54$ \\
\bottomrule
\end{tabular}
\end{table}

Across the $5{\times}3$ model/dataset grid, $87.5\%$ of layer assignments remain unchanged across calibration tasks, with cross-task Spearman correlation $\rho{=}0.757$. The layers that change are mostly near the first and last transformer blocks, indicating that many high-precision assignments are shared across tasks while task-specific differences concentrate in a smaller set of boundary layers.

\input{figs/mixed_task_heatmap}

\subsection{Calibration source vs.\ allocation policy}
\label{app:calib_vs_policy}

This subsection consolidates the controlled ablation used to disentangle the contribution of \emph{task-aligned calibration data} from the contribution of the \emph{task-aware allocation policy}. We pose two sub-questions:
(\textbf{AQ1}) If we perfectly align calibration data with the test task (on-task calibration), can a task-agnostic policy (AWQ) recover full-precision performance?
(\textbf{AQ2}) Does a task-aware allocation policy provide robustness that data alignment alone cannot, especially in regimes where standard minimization is unstable?

We compare three setups on TriviaQA across an auxiliary AWQ-supported model suite:
(i) \textbf{AWQ-Mixed}: balanced pool of math, code, and trivia ($N{=}2048$);
(ii) \textbf{AWQ-OT}: $512$ examples drawn exclusively from TriviaQA (perfect on-task alignment, tests AQ1);
(iii) \textbf{TAQ-IS / TAQ-O}: our task-aware policies on TriviaQA. We use this auxiliary suite because GPTQ checkpoints were unavailable for some main-experiment models; running AWQ/AWQ-OT under the same PTQ codepath gives a consistent comparison. Since this ablation uses a different sampled setting from the main TriviaQA experiments ($512$ samples), absolute values may differ slightly from \autoref{tab:trivia_appx}.

The experiment refutes \textbf{AQ1} and supports \textbf{AQ2}. Under \textbf{AQ1}, AWQ-OT should consistently improve over AWQ-Mixed, but its behavior is unstable: Llama-3.1 improves ($53.37 \!\to\! 55.66$ EM) while Phi-4 collapses ($2.25 \!\to\! 0.73$ EM) and Mistral changes minimally. Under \textbf{AQ2}, the best TAQ policy is more stable across this auxiliary suite, nearly matching FP16 on Phi-4 with TAQ-IS ($42.33$ vs.\ $43.16$ EM) and exceeding AWQ-OT on Qwen3 with TAQ-IS ($11.82$ vs.\ $9.81$ EM), while TAQ-O gives the highest Qwen3 EM ($18.51$).

\begin{table*}[H]
\centering
\caption{\textbf{Component analysis (calibration data vs.\ policy)} on TriviaQA across an auxiliary model suite. \textbf{AWQ (Mixed)} uses a mixed calibration pool; \textbf{AWQ-OT} uses on-task TriviaQA calibration; \textbf{TAQ-IS / TAQ-O} are our task-aware policies. \textbf{Bold} marks the best quantized result per metric/backbone; FP16 is shown as a full-precision reference.}
\label{tab:component_analysis}
\small
\setlength{\tabcolsep}{5pt}
\begin{tabular}{l cc cc cc cc cc}
\toprule
\multirow{2}{*}{\textbf{Method}}
& \multicolumn{2}{c}{\textbf{Phi-4}}
& \multicolumn{2}{c}{\textbf{Qwen3}}
& \multicolumn{2}{c}{\textbf{Llama-3.1}}
& \multicolumn{2}{c}{\textbf{Qwen2.5}}
& \multicolumn{2}{c}{\textbf{Mistral}} \\
\cmidrule(lr){2-3}\cmidrule(lr){4-5}\cmidrule(lr){6-7}\cmidrule(lr){8-9}\cmidrule(lr){10-11}
& EM & F1 & EM & F1 & EM & F1 & EM & F1 & EM & F1 \\
\midrule
W\textbackslash O (FP16) & 43.16 & 51.24 & 11.23 & 26.33 & 57.86 & 66.00 & 11.43 & 21.90 & 18.90 & 35.18 \\
\midrule
AWQ (Mixed)      &  2.25 &  7.07 &  9.23 & 23.94 & 53.37 & 62.23 & 11.91 & 24.06 & 17.48 & 33.75 \\
AWQ-OT (On-Task) &  0.73 &  4.54 &  9.81 & 24.33 & 55.66 & 62.92 & 10.59 & 24.21 & 17.58 & 26.33 \\
\midrule
\textbf{TAQ-IS (Ours)} & \textbf{42.33} & \textbf{50.81} & 11.82 & 26.12 & 57.03 & \textbf{65.35} & 12.06 & 22.47 & \textbf{18.75} & \textbf{34.74} \\
\textbf{TAQ-O  (Ours)} &  0.73 &  4.57 & \textbf{18.51} & \textbf{34.34} & \textbf{57.91} & 65.14 & \textbf{12.65} & \textbf{25.49} & 18.51 & 34.34 \\
\bottomrule
\end{tabular}
\end{table*}

\section{Hardware and deployment}
\label{app:hardware}

We benchmarked all methods on NVIDIA A40 hardware across three models, two batch sizes, and with GPTQ-Int4 and AWQ-Int4 baselines from HuggingFace, for a total of $45$ configurations.

\subsection{Throughput and latency}

The headline results for Qwen2.5--7B are shown in \autoref{tab:qwen25-7b-benchmarks}. TAQ-IS achieves a $35\%$ throughput improvement at batch size~$1$---a measured speedup on A40 hardware, not a theoretical estimate. At batch size~$8$, the gain persists at $+11.6\%$ in the full benchmark sweep. For Qwen2--7B against an FP32 baseline, TAQ-O delivers a $4.56\times$ throughput improvement, increasing throughput from $8.0$ to $36.3$ tokens/s in the same sweep.

\begin{table}[H]
\centering
\caption{Real hardware benchmark results for Qwen2.5--7B on NVIDIA A40 (batch size~1; throughput in tokens/s, latency in ms/token).}
\label{tab:qwen25-7b-benchmarks}
\begin{tabular}{lccc}
\toprule
\textbf{Method} & \textbf{Throughput$\uparrow$} & \textbf{Latency$\downarrow$} & \textbf{vs.\ FP16} \\
\midrule
FP16        & 27.63 & 36.25  & baseline \\
TAQ-IS      & 37.29 & 26.82  & $+35\%$ / $-26\%$ \\
TAQ-KL      & 32.83 & 30.48  & $+19\%$ / $-16\%$ \\
TAQ-O       & 27.86 & 35.93  & $+1\%$  / $-1\%$  \\
UNIFORM4    & 28.71 & 34.83  & $+4\%$  / $-4\%$  \\
GPTQ-Int4   &  4.40 & 227.26 & $-84\%$ / $+527\%$ \\
AWQ-Int4    & 16.12 &  62.51 & $-42\%$ / $+72\%$ \\
\bottomrule
\end{tabular}
\end{table}

\subsection{Memory accounting}
\label{app:memory_accounting}

We compare TAQ's realized weight footprint against uniform 4-bit GPTQ and AWQ under a single convention: $W$ counts the stored weight tensors needed for inference, including quantized linear weights, FP16 group-wise scales with group size $G{=}128$, per-group zero-points at the layer's bitwidth, FP16-retained linear blocks (TAQ-O only), input embeddings and \texttt{lm\_head} in FP16 (tied where the model ties them), and final/RMS layer norms in FP16. Concretely, with $b_\ell \in \{4,8,16\}$, per-layer transformer-linear parameter count $N_\ell$, vocabulary size $V$, hidden size $h$, and $\mathbf{1}_{\mathrm{untied}}$ indicating untied embeddings,
We report $W = W_{\mathrm{bytes}}/2^{30}$ in GB. Under this convention, our GPTQ/AWQ numbers reproduce the realized HuggingFace 4-bit checkpoints to within ${<}2\%$; for example, Qwen2.5--7B gives $5.19$~GB.

\begin{table}[H]
\centering
\caption{Realized weight footprint for Qwen2.5--7B .}
\label{tab:gptq-awq-comparison}
\begin{tabular}{lccc}
\toprule
\textbf{Method} & \textbf{Avg.\ bits/param} & \textbf{Realized $W$} & \textbf{Allocation strategy} \\
\midrule
FP16 baseline    & 16.0 & 14.18 GB & none \\
GPTQ / AWQ       &  4.0 &  5.19 GB & uniform 4-bit \\
TAQ-IS / TAQ-KL  &  5.0 &  5.95 GB & task-aware mixed 4/8-bit \\
\bottomrule
\end{tabular}
\end{table}

The $0.76$~GB premium over uniform 4-bit GPTQ/AWQ is consistent with the $5$-vs-$4$ average bit budget; it buys 8-bit precision on the critical $25\%$ of layers identified by our task-aware allocation. The component breakdown for TAQ-IS / TAQ-KL is $3.80$~GB quantized linears $+$ $0.10$~GB scales $+$ $0.03$~GB zero-points $+$ $2.03$~GB FP16 embeddings/\texttt{lm\_head} $+$ ${<}0.01$~GB norms.

\subsection{Real-kernel validation}
\label{app:real_kernels}

To demonstrate that TAQ's per-layer bit allocation is deployable with real quantized kernels, we replaced each layer with hardware-accelerated equivalents following TAQ's \texttt{bit\_map}: NF4 ($4$-bit) kernels for layers assigned $4$-bit, and Int8 kernels for layers assigned $8$-bit. Results for Qwen2.5--3B are shown in \autoref{tab:real-kernel-validation}.

\begin{table}[H]
\centering
\caption{Real-kernel memory validation on Qwen2.5--3B.}
\label{tab:real-kernel-validation}
\begin{tabular}{lccc}
\toprule
\textbf{Method} & \textbf{Weight size} & \textbf{Peak memory} & \textbf{Compression vs.\ FP16} \\
\midrule
FP16 baseline     & 5.79 GB & 5.79 GB & $1.0\times$ \\
GPTQ-Int4         & 1.93 GB & 2.06 GB & $3.0\times$ \\
AWQ-Int4          & 1.92 GB & 1.96 GB & $3.0\times$ \\
TAQ real kernels  & 2.20 GB & 2.31 GB & $2.6\times$ \\
\bottomrule
\end{tabular}
\end{table}

TAQ with real kernels achieves $2.6\times$ memory compression with working inference. The additional $0.27$--$0.28$~GB over GPTQ/AWQ corresponds to the $25\%$ of layers kept at $8$-bit precision---the layers identified by our analysis as critical for task performance.

\section{Inter-layer analyses}
\label{app:interlayer}

This section consolidates the analyses referenced from \autoref{exp:inter_layer_analysis}. \autoref{app:interlayer_stability} reports per-architecture aggregate statistics, per-layer Spearman correlations, scatter plots, and the cosine-similarity decay vs.\ distance from the cutoff layer for the inter-layer score-stability experiment. 

\subsection{Inter-layer score stability under upstream quantization}
\label{app:interlayer_stability}

\begin{figure}[H]
\centering
\includegraphics[width=\linewidth]{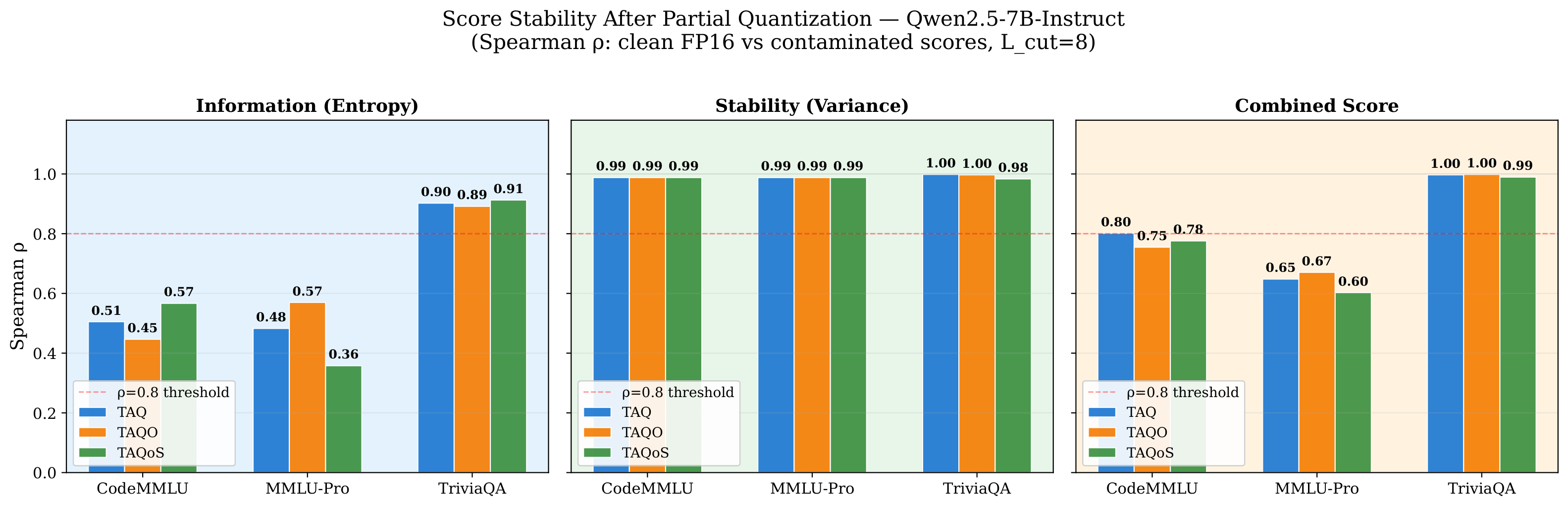}
\caption{\textbf{Clean-vs-contaminated layer-score stability on Qwen2.5-7B.} Spearman correlations between clean FP16 layer-importance scores and scores recomputed after quantizing the first eight transformer blocks. The full combined score is highly stable on TriviaQA and moderately stable on CodeMMLU and MMLU-Pro, while the variance/stability signal is near-perfect across tasks.}
\label{fig:il_spearman_qwen}
\end{figure}

\begin{figure}[H]
\centering
\includegraphics[width=\linewidth]{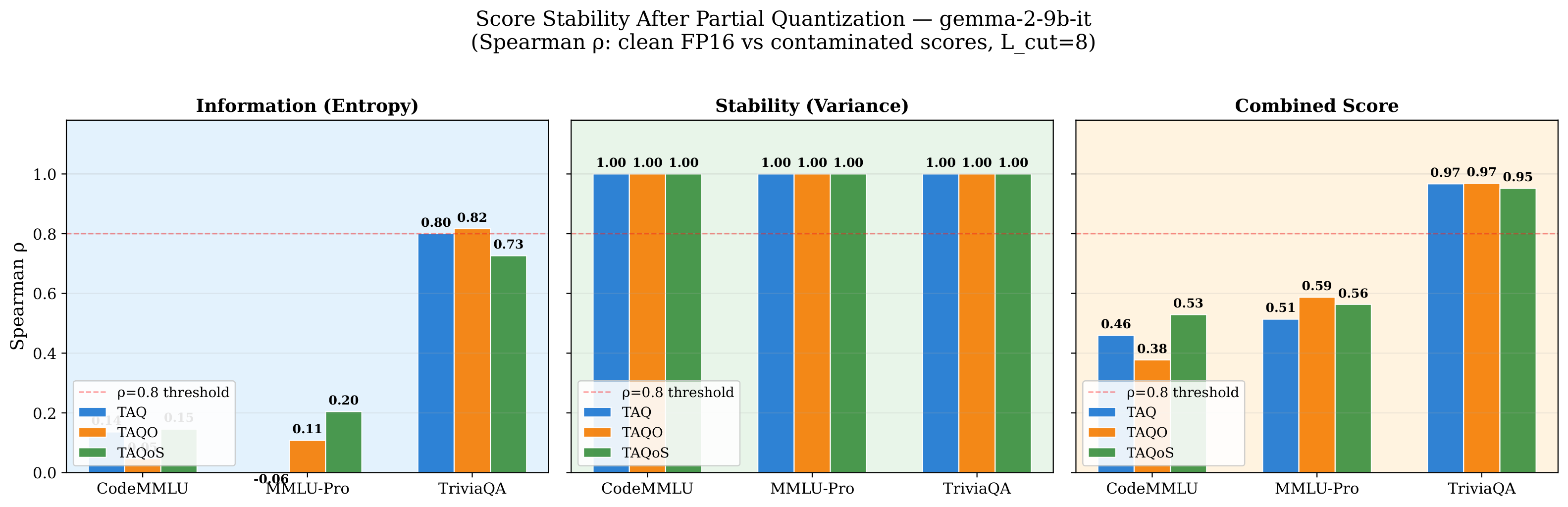}
\caption{\textbf{Clean-vs-contaminated(quantized) first 8 layers-score stability on Gemma-2-9B.} }
\label{fig:il_spearman_gemma}
\end{figure}

\begin{figure}[H]
\centering
\includegraphics[width=\linewidth]{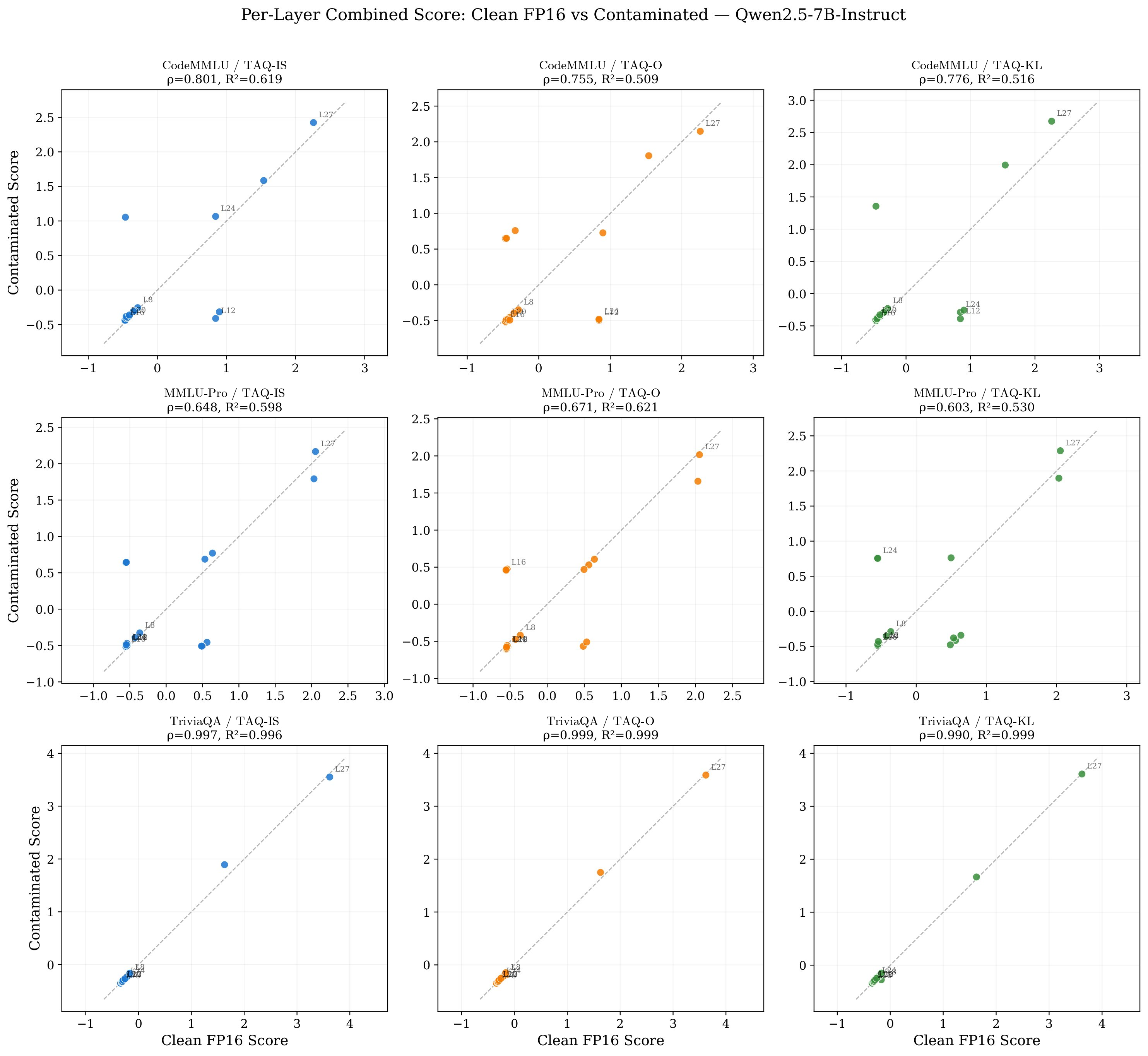}
\caption{\textbf{Per-layer clean-vs-contaminated score agreement on Qwen2.5-7B.} Each panel compares clean FP16 layer-importance scores against scores obtained after upstream 4-bit quantization of the first eight transformer blocks. Alignment is strongest on TriviaQA, supporting clean FP16 scores as useful proxies rather than exact joint-quantization scores.}
\label{fig:il_scatter_qwen}
\end{figure}

\begin{figure}[H]
\centering
\includegraphics[width=\linewidth]{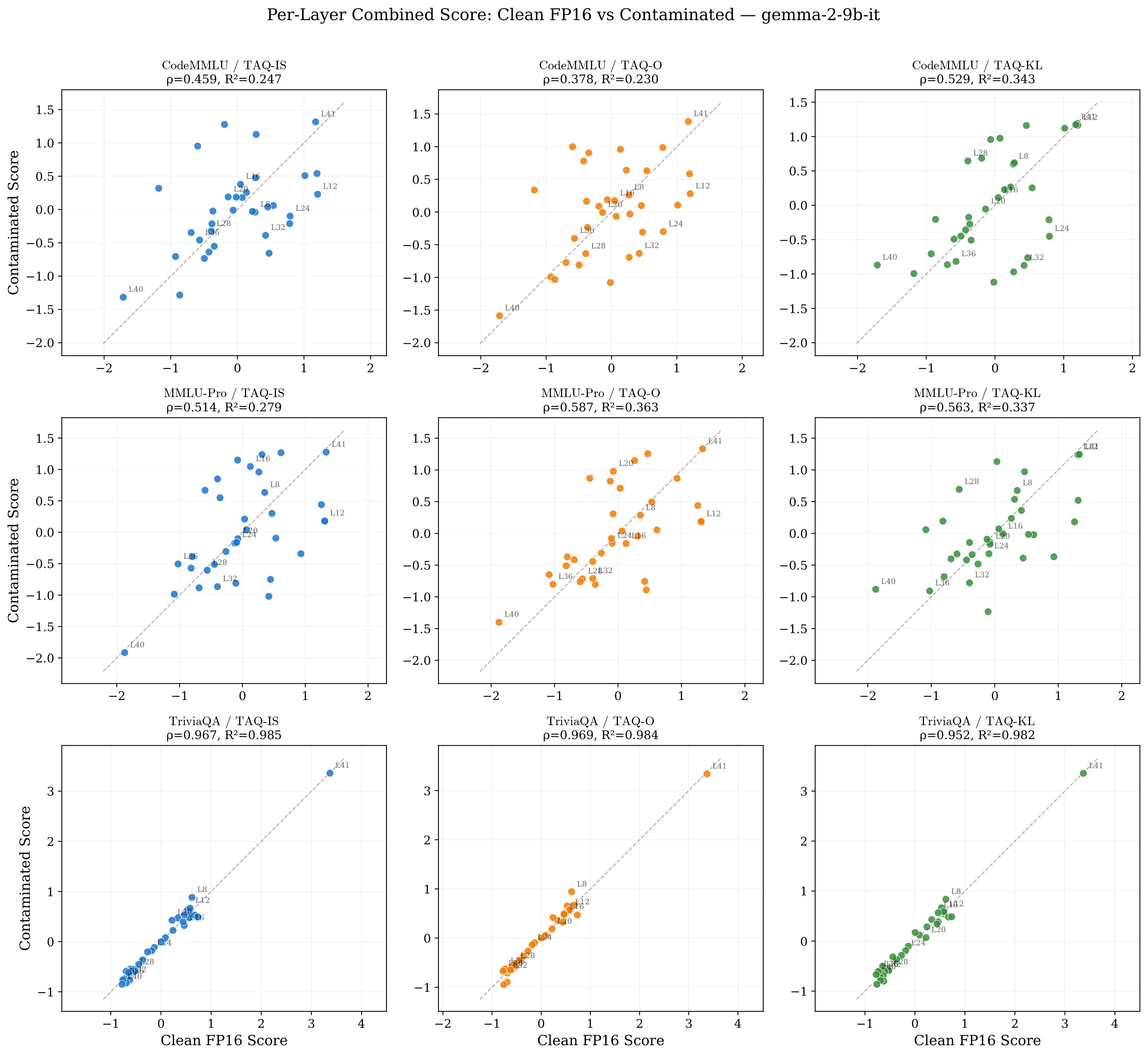}
\caption{\textbf{Per-layer clean-vs-contaminated score agreement on Gemma-2-9B.} TriviaQA remains highly stable, while CodeMMLU and MMLU-Pro show larger deviations, suggesting stronger task-dependent inter-layer interactions.}
\label{fig:il_scatter_gemma}
\end{figure}

\autoref{tab:appx_contamination_arch} reports per-architecture aggregate statistics for the inter-layer score-stability experiment after quantizing the first eight transformer blocks. The stability-signal Spearman column corresponds to the variance/stability signal, while the combined-score column corresponds to the full TAQ score. \autoref{fig:il_spearman_qwen} and \autoref{fig:il_spearman_gemma} report per-task Spearman correlations between clean FP16 layer-importance rankings and rankings recomputed after quantizing the first eight transformer blocks. \autoref{fig:il_scatter_qwen} and \autoref{fig:il_scatter_gemma} provide per-layer scatter plots, while \autoref{tab:appx_contamination_distance} reports the cosine-similarity decay vs.\ distance from the cutoff layer for the canonical Qwen2.5-7B / TriviaQA / TAQ-IS run. Complementary cross-model summary, propagation, and per-layer activation MSE figures are provided as \autoref{fig:appx_interlayer_xmodel}, \autoref{fig:appx_il_propagation}, and \autoref{fig:appx_perlayer_mse}.

\begin{table}[H]
\centering
\caption{\textbf{Per-architecture aggregate statistics} of the inter-layer score-stability experiment (first $8$ blocks quantized, downstream layers re-scored against FP16). Values are mean $\pm$ standard deviation across datasets and scoring-family runs where applicable. The stability-signal column corresponds to the variance/stability signal; the combined-score column corresponds to the full TAQ score.}
\label{tab:appx_contamination_arch}
\small
\setlength{\tabcolsep}{8pt}
\begin{tabular}{l ccc}
\toprule
\textbf{Architecture} & \textbf{Stability-score $\rho$} & \textbf{Combined-score $\rho$} & \textbf{FP16 cosine sim} \\
\midrule
Qwen2.5-7B & $0.990 \pm 0.005$ & $0.804 \pm 0.156$ & $0.931 \pm 0.003$ \\
Gemma-2-9B & $1.000 \pm 0.000$ & $0.658 \pm 0.237$ & $0.970 \pm 0.006$ \\
\bottomrule
\end{tabular}
\end{table}

\begin{table}[H]
\centering
\caption{\textbf{Cosine-similarity decay vs.\ distance from cutoff} on Qwen2.5-7B / TriviaQA / TAQ-IS (first $8$ blocks quantized). Cosine similarity drops sharply at distance $+1$ and then stabilizes in the $0.93$--$0.95$ range, consistent with a stable residual-stream propagation plateau.}
\label{tab:appx_contamination_distance}
\small
\setlength{\tabcolsep}{8pt}
\begin{tabular}{c c l}
\toprule
\textbf{Distance} & \textbf{Cosine sim} & \textbf{Interpretation} \\
\midrule
$0$  (layer $8$)   & 1.000 & Cutoff layer (no error yet) \\
$+1$ (layer $9$)   & 0.880 & Sharp initial drop \\
$+7$ (layer $15$)  & 0.932 & Stabilization \\
$+19$ (layer $27$) & 0.933 & Sustained plateau \\
\bottomrule
\end{tabular}
\end{table}

%
%
\input{figs/interlayer_crossmodel}
\input{figs/il_propagation}
\input{figs/perlayer_mse}

%% file: figs/mixed_task_heatmap.tex
\begin{figure}[H]
\centering
\includegraphics[width=0.95\linewidth]{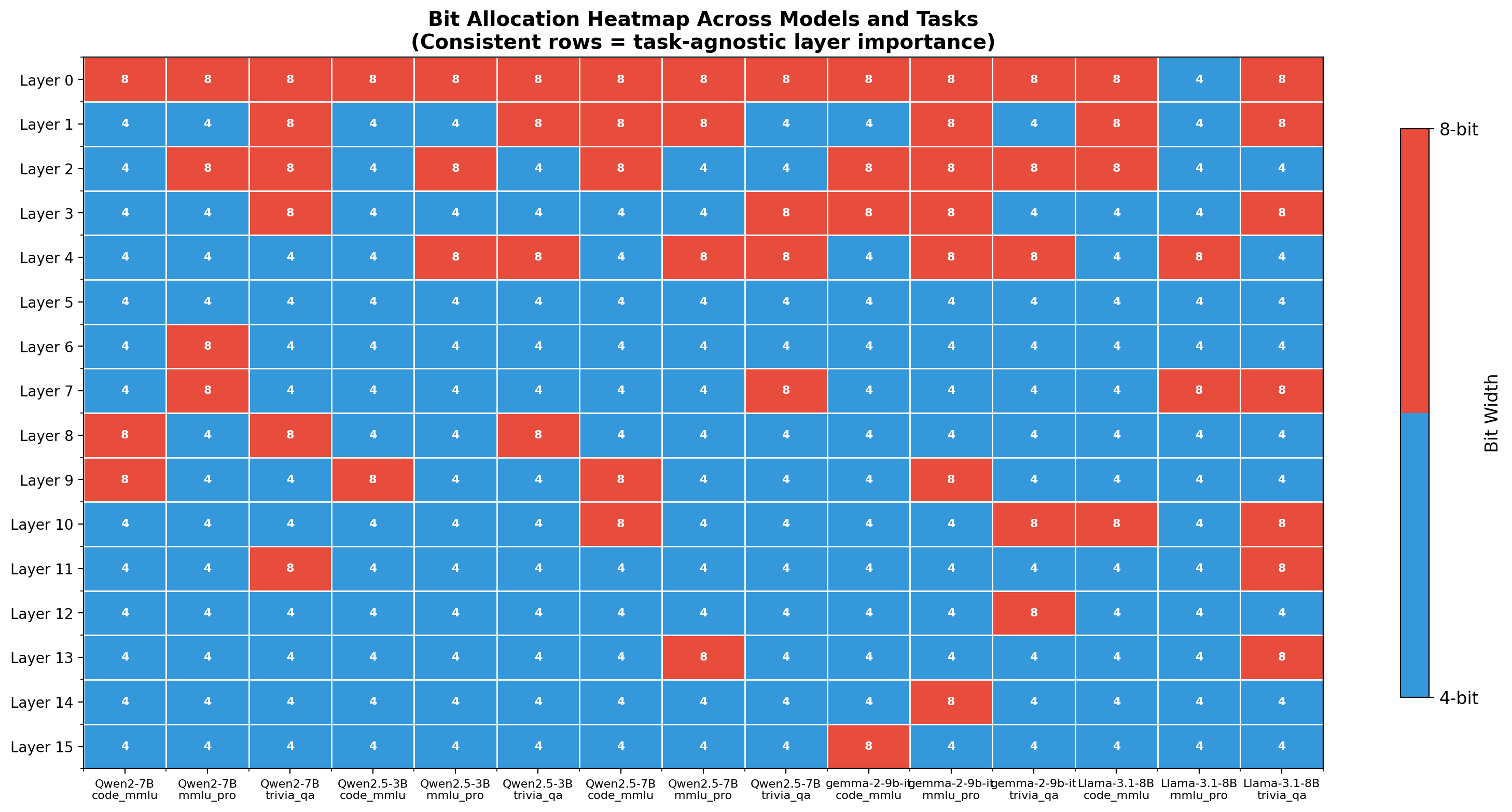}
\caption{\textbf{Bit-allocation heatmap across $5$ backbones $\times$ $3$ datasets.} $87.5\%$ of layers receive identical bit-widths regardless of calibration task; task-specific variation concentrates at boundary blocks.}
\label{fig:appx_mixed_heatmap}
\end{figure}

%% file: figs/interlayer_crossmodel.tex
\begin{figure}[H]
\centering
\includegraphics[width=0.95\linewidth]{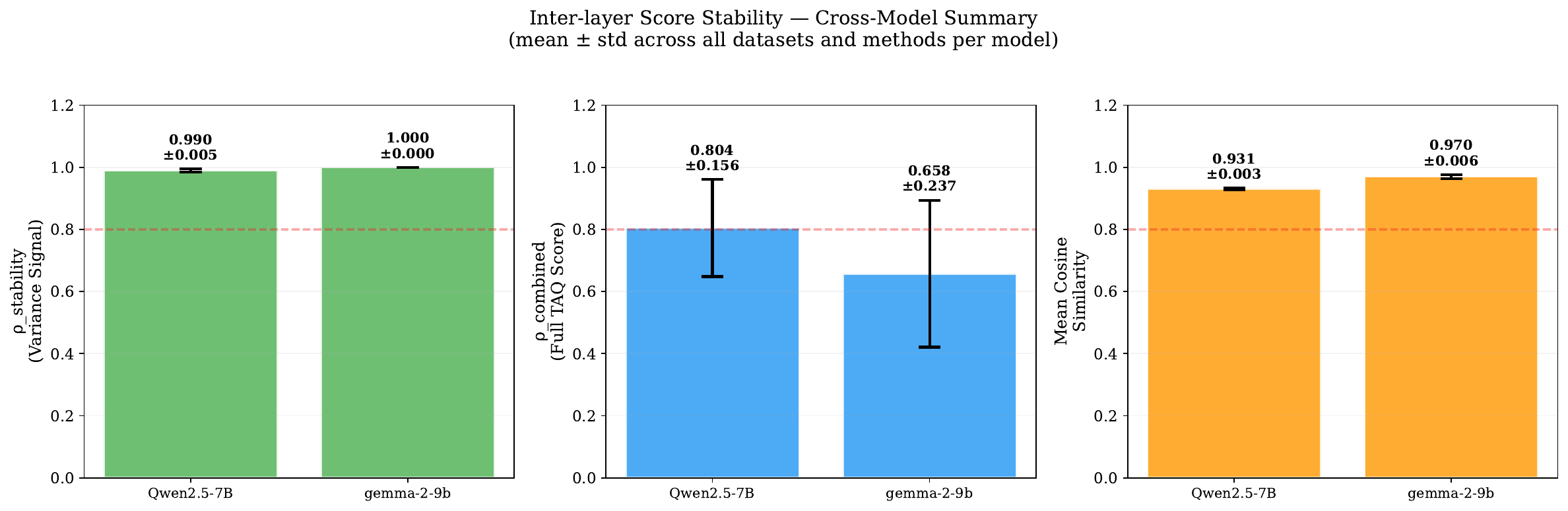}
\caption{\textbf{Cross-model summary of the inter-layer contamination experiment.} Rank-preservation Spearman $\rho$, FP16 cosine similarity, and activation MSE aggregated over both architectures and the three TAQ families; the residual-stream propagation pattern is consistent across models.}
\label{fig:appx_interlayer_xmodel}
\end{figure}

%% file: figs/il_propagation.tex
\begin{figure}[H]
\centering

\includegraphics[width=\linewidth]{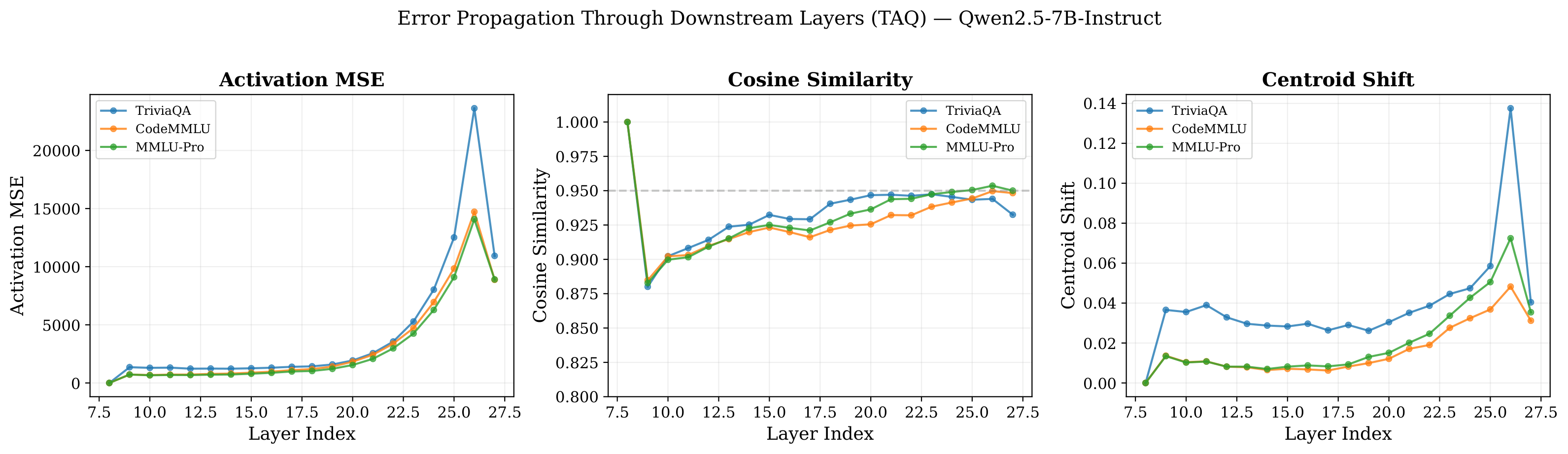}

\includegraphics[width=\linewidth]{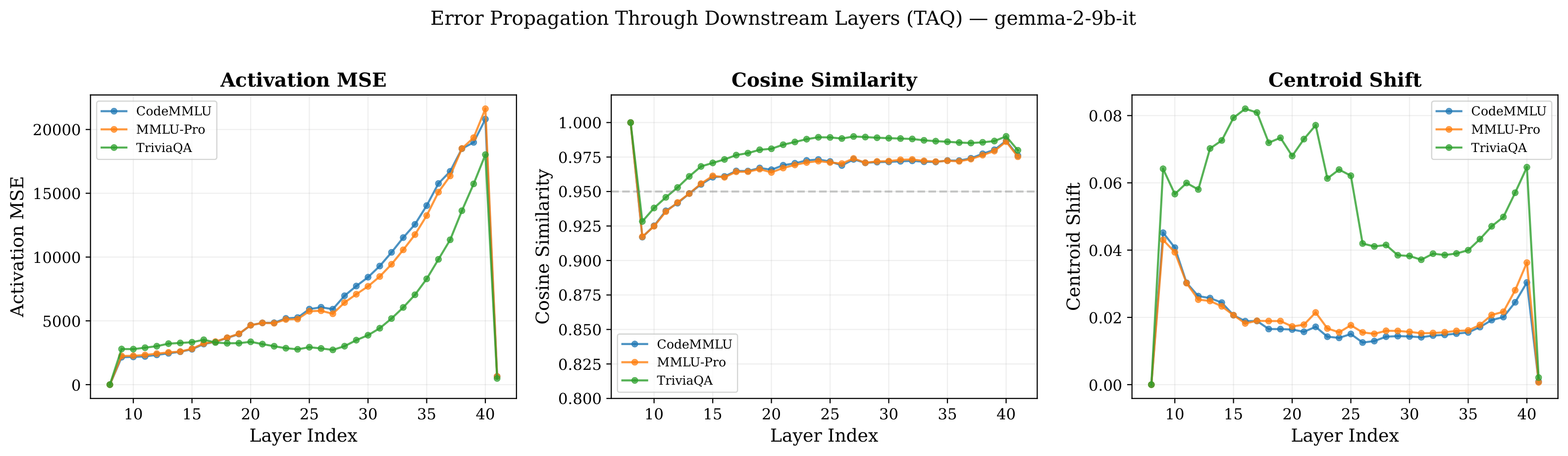}

\caption{\textbf{Error-propagation curves.} Cosine similarity vs.\ distance from the cutoff, illustrating the residual-stream propagation pattern on Qwen2.5-7B and Gemma-2-9B for TAQ-IS.}
\label{fig:appx_il_propagation}
\end{figure}

%% file: figs/perlayer_mse.tex
\begin{figure}[H]
\centering
\includegraphics[width=0.95\linewidth]{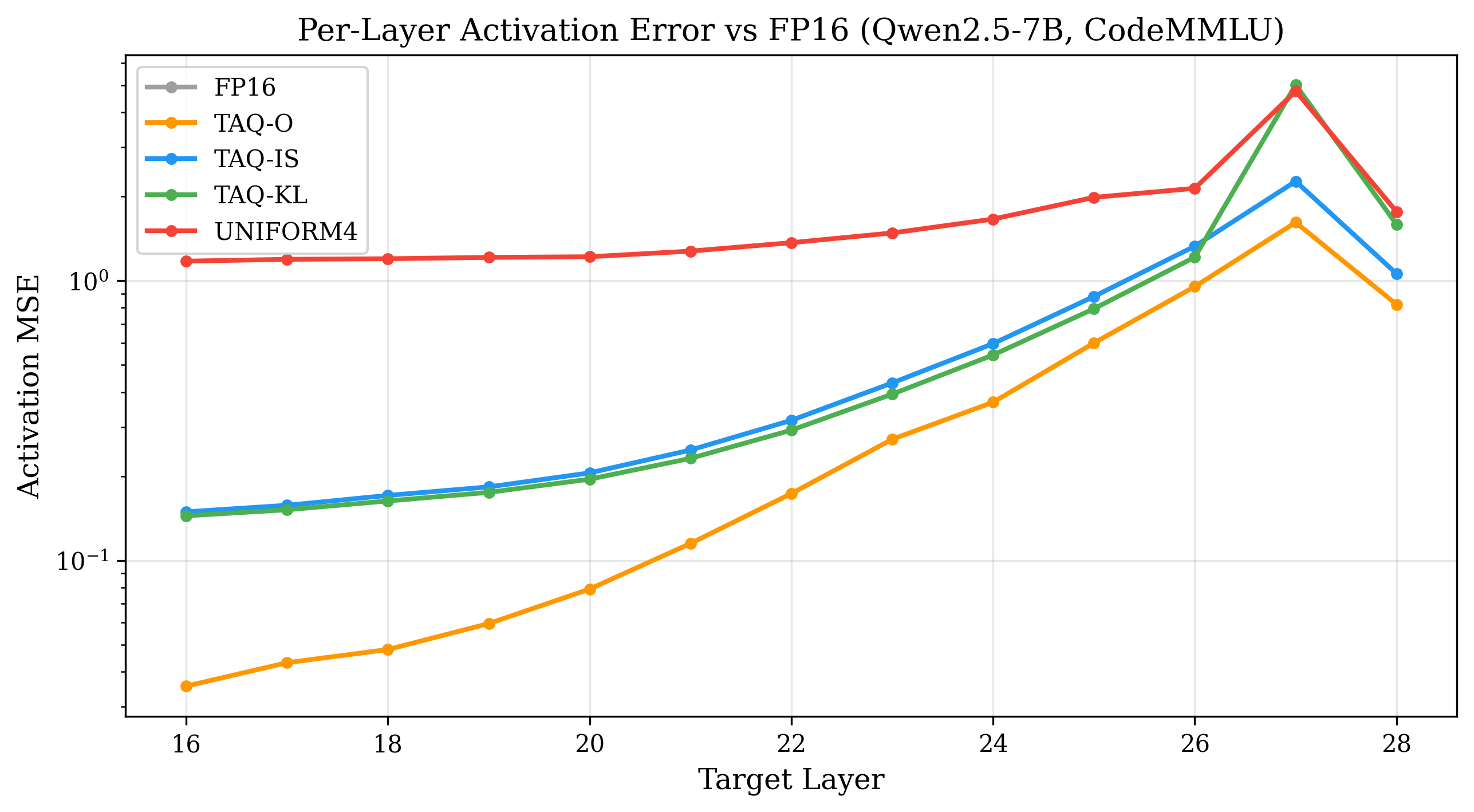}
\caption{\textbf{Per-layer activation MSE} for the three TAQ variants and a uniform 4-bit reference; distortion concentrates near boundary blocks.}
\label{fig:appx_perlayer_mse}
\end{figure}

%% file: checklist.tex
\section*{NeurIPS Paper Checklist}

\begin{enumerate}

\item {\bf Claims}
    \item[] Question: Do the main claims made in the abstract and introduction accurately reflect the paper's contributions and scope?
    \item[] Answer: \answerYes{}.
    \item[] Justification: The abstract and introduction clearly state TAQ's contributions, assumptions, and scope, including that it is a training-free weight-only precision-allocation policy and that TAQ-O is diagnostic. The claims align with the stated empirical, hardware, robustness, and residual-stream analyses.
    \item[] Guidelines:
    \begin{itemize}
        \item The answer \answerNA{} means that the abstract and introduction do not include the claims made in the paper.
        \item The abstract and/or introduction should clearly state the claims made, including the contributions made in the paper and important assumptions and limitations. A \answerNo{} or \answerNA{} answer to this question will not be perceived well by the reviewers. 
        \item The claims made should match theoretical and experimental results, and reflect how much the results can be expected to generalize to other settings. 
        \item It is fine to include aspirational goals as motivation as long as it is clear that these goals are not attained by the paper. 
    \end{itemize}

\item {\bf Limitations}
    \item[] Question: Does the paper discuss the limitations of the work performed by the authors?
    \item[] Answer: \answerYes{} 
    \item[] Justification: The paper includes a dedicated \textbf{Limitations.} paragraph discussing dependence on representative calibration prompts, possible degradation under distribution shift, coarse layer-level precision allocation, proxy-based allocation limitations, and hardware-dependent practical gains.
    \item[] Guidelines:
    \begin{itemize}
        \item The answer \answerNA{} means that the paper has no limitation while the answer \answerNo{} means that the paper has limitations, but those are not discussed in the paper. 
        \item The authors are encouraged to create a separate ``Limitations'' section in their paper.
        \item The paper should point out any strong assumptions and how robust the results are to violations of these assumptions (e.g., independence assumptions, noiseless settings, model well-specification, asymptotic approximations only holding locally). The authors should reflect on how these assumptions might be violated in practice and what the implications would be.
        \item The authors should reflect on the scope of the claims made, e.g., if the approach was only tested on a few datasets or with a few runs. In general, empirical results often depend on implicit assumptions, which should be articulated.
        \item The authors should reflect on the factors that influence the performance of the approach. For example, a facial recognition algorithm may perform poorly when image resolution is low or images are taken in low lighting. Or a speech-to-text system might not be used reliably to provide closed captions for online lectures because it fails to handle technical jargon.
        \item The authors should discuss the computational efficiency of the proposed algorithms and how they scale with dataset size.
        \item If applicable, the authors should discuss possible limitations of their approach to address problems of privacy and fairness.
        \item While the authors might fear that complete honesty about limitations might be used by reviewers as grounds for rejection, a worse outcome might be that reviewers discover limitations that aren't acknowledged in the paper. The authors should use their best judgment and recognize that individual actions in favor of transparency play an important role in developing norms that preserve the integrity of the community. Reviewers will be specifically instructed to not penalize honesty concerning limitations.
    \end{itemize}

\item {\bf Theory assumptions and proofs}
    \item[] Question: For each theoretical result, does the paper provide the full set of assumptions and a complete (and correct) proof?
    \item[] Answer: \answerNA{} 
    \item[] Justification: The paper contains methodological objectives and scoring/allocation formulas, but it does not state formal theoretical results, theorems, or lemmas requiring assumptions and proofs.
    \item[] Guidelines:
    \begin{itemize}
        \item The answer \answerNA{} means that the paper does not include theoretical results. 
        \item All the theorems, formulas, and proofs in the paper should be numbered and cross-referenced.
        \item All assumptions should be clearly stated or referenced in the statement of any theorems.
        \item The proofs can either appear in the main paper or the supplemental material, but if they appear in the supplemental material, the authors are encouraged to provide a short proof sketch to provide intuition. 
        \item Inversely, any informal proof provided in the core of the paper should be complemented by formal proofs provided in appendix or supplemental material.
        \item Theorems and Lemmas that the proof relies upon should be properly referenced. 
    \end{itemize}

    \item {\bf Experimental result reproducibility}
    \item[] Question: Does the paper fully disclose all the information needed to reproduce the main experimental results of the paper to the extent that it affects the main claims and/or conclusions of the paper (regardless of whether the code and data are provided or not)?
    \item[] Answer: \answerYes{} 
    \item[] Justification: The paper discloses the main experimental setup, including datasets, metrics, model backbones, baselines, quantization protocol, calibration/test splits, memory accounting, and TAQ hyperparameters, and provides a reference implementation for reproduction.
    \item[] Guidelines:
    \begin{itemize}
        \item The answer \answerNA{} means that the paper does not include experiments.
        \item If the paper includes experiments, a \answerNo{} answer to this question will not be perceived well by the reviewers: Making the paper reproducible is important, regardless of whether the code and data are provided or not.
        \item If the contribution is a dataset and\slash or model, the authors should describe the steps taken to make their results reproducible or verifiable. 
        \item Depending on the contribution, reproducibility can be accomplished in various ways. For example, if the contribution is a novel architecture, describing the architecture fully might suffice, or if the contribution is a specific model and empirical evaluation, it may be necessary to either make it possible for others to replicate the model with the same dataset, or provide access to the model. In general. releasing code and data is often one good way to accomplish this, but reproducibility can also be provided via detailed instructions for how to replicate the results, access to a hosted model (e.g., in the case of a large language model), releasing of a model checkpoint, or other means that are appropriate to the research performed.
        \item While NeurIPS does not require releasing code, the conference does require all submissions to provide some reasonable avenue for reproducibility, which may depend on the nature of the contribution. For example
        \begin{enumerate}
            \item If the contribution is primarily a new algorithm, the paper should make it clear how to reproduce that algorithm.
            \item If the contribution is primarily a new model architecture, the paper should describe the architecture clearly and fully.
            \item If the contribution is a new model (e.g., a large language model), then there should either be a way to access this model for reproducing the results or a way to reproduce the model (e.g., with an open-source dataset or instructions for how to construct the dataset).
            \item We recognize that reproducibility may be tricky in some cases, in which case authors are welcome to describe the particular way they provide for reproducibility. In the case of closed-source models, it may be that access to the model is limited in some way (e.g., to registered users), but it should be possible for other researchers to have some path to reproducing or verifying the results.
        \end{enumerate}
    \end{itemize}

\item {\bf Open access to data and code}
    \item[] Question: Does the paper provide open access to the data and code, with sufficient instructions to faithfully reproduce the main experimental results, as described in supplemental material?
    \item[] Answer: \answerYes{} 
    \item[] Justification: The paper provides an anonymized reference implementation link and specifies the public benchmarks, calibration/evaluation splits, and quantization settings used for the main experiments.
    \item[] Guidelines:
    \begin{itemize}
        \item The answer \answerNA{} means that paper does not include experiments requiring code.
        \item Please see the NeurIPS code and data submission guidelines (\url{https://neurips.cc/public/guides/CodeSubmissionPolicy}) for more details.
        \item While we encourage the release of code and data, we understand that this might not be possible, so \answerNo{} is an acceptable answer. Papers cannot be rejected simply for not including code, unless this is central to the contribution (e.g., for a new open-source benchmark).
        \item The instructions should contain the exact command and environment needed to run to reproduce the results. See the NeurIPS code and data submission guidelines (\url{https://neurips.cc/public/guides/CodeSubmissionPolicy}) for more details.
        \item The authors should provide instructions on data access and preparation, including how to access the raw data, preprocessed data, intermediate data, and generated data, etc.
        \item The authors should provide scripts to reproduce all experimental results for the new proposed method and baselines. If only a subset of experiments are reproducible, they should state which ones are omitted from the script and why.
        \item At submission time, to preserve anonymity, the authors should release anonymized versions (if applicable).
        \item Providing as much information as possible in supplemental material (appended to the paper) is recommended, but including URLs to data and code is permitted.
    \end{itemize}

\item {\bf Experimental setting/details}
    \item[] Question: Does the paper specify all the training and test details (e.g., data splits, hyperparameters, how they were chosen, type of optimizer) necessary to understand the results?
    \item[] Answer: \answerYes{} 
    \item[] Justification: \autoref{sec:settings} specifies the tasks, metrics, models, baselines, quantization protocol, calibration/test splits, fixed sampling, and key TAQ hyperparameters; no optimizer details are needed because the method is training-free and applies no gradient-based updates.
    \item[] Guidelines:
    \begin{itemize}
        \item The answer \answerNA{} means that the paper does not include experiments.
        \item The experimental setting should be presented in the core of the paper to a level of detail that is necessary to appreciate the results and make sense of them.
        \item The full details can be provided either with the code, in appendix, or as supplemental material.
    \end{itemize}

\item {\bf Experiment statistical significance}
    \item[] Question: Does the paper report error bars suitably and correctly defined or other appropriate information about the statistical significance of the experiments?
    \item[] Answer: \answerYes{} 
    \item[] Justification: \autoref{exp:inter_layer_analysis} reports Spearman's $\rho$ between clean FP16 and post-quantization layer rankings, with 1-sigma std across the $3{\times}3$ task$\times$scoring grid per architecture (\autoref{tab:appx_contamination_arch}). Spearman is non-parametric and matches TAQ's rank-based layer-selection objective. Main accuracy and hardware numbers are fixed-protocol point estimates since TAQ is training-free with fixed splits and no training variance.

    \item[] Guidelines:
    \begin{itemize}
        \item The answer \answerNA{} means that the paper does not include experiments.
        \item The authors should answer \answerYes{} if the results are accompanied by error bars, confidence intervals, or statistical significance tests, at least for the experiments that support the main claims of the paper.
        \item The factors of variability that the error bars are capturing should be clearly stated (for example, train/test split, initialization, random drawing of some parameter, or overall run with given experimental conditions).
        \item The method for calculating the error bars should be explained (closed form formula, call to a library function, bootstrap, etc.)
        \item The assumptions made should be given (e.g., Normally distributed errors).
        \item It should be clear whether the error bar is the standard deviation or the standard error of the mean.
        \item It is OK to report 1-sigma error bars, but one should state it. The authors should preferably report a 2-sigma error bar than state that they have a 96\% CI, if the hypothesis of Normality of errors is not verified.
        \item For asymmetric distributions, the authors should be careful not to show in tables or figures symmetric error bars that would yield results that are out of range (e.g., negative error rates).
        \item If error bars are reported in tables or plots, the authors should explain in the text how they were calculated and reference the corresponding figures or tables in the text.
    \end{itemize}

\item {\bf Experiments compute resources}
    \item[] Question: For each experiment, does the paper provide sufficient information on the computer resources (type of compute workers, memory, time of execution) needed to reproduce the experiments?
    \item[] Answer: \answerYes{} 
    \item[] Justification:  The paper provides compute-resource details in subsection~\ref{subsec:compute_resources}, including the single-GPU NVIDIA A40 workers with 48 GB memory, approximate per-run GPU-hour costs, total reported compute, and total project compute including preliminary and failed runs.
    \item[] Guidelines:
    \begin{itemize}
        \item The answer \answerNA{} means that the paper does not include experiments.
        \item The paper should indicate the type of compute workers CPU or GPU, internal cluster, or cloud provider, including relevant memory and storage.
        \item The paper should provide the amount of compute required for each of the individual experimental runs as well as estimate the total compute. 
        \item The paper should disclose whether the full research project required more compute than the experiments reported in the paper (e.g., preliminary or failed experiments that didn't make it into the paper). 
    \end{itemize}
    
\item {\bf Code of ethics}
    \item[] Question: Does the research conducted in the paper conform, in every respect, with the NeurIPS Code of Ethics \url{https://neurips.cc/public/EthicsGuidelines}?
    \item[] Answer: \answerYes{} 
    \item[] Justification: The authors have reviewed the NeurIPS Code of Ethics, and the work conforms to it: the paper studies training-free LLM quantization using existing benchmarks and open-weight models, with no human-subject experiments or new sensitive-data collection.
    \item[] Guidelines:
    \begin{itemize}
        \item The answer \answerNA{} means that the authors have not reviewed the NeurIPS Code of Ethics.
        \item If the authors answer \answerNo, they should explain the special circumstances that require a deviation from the Code of Ethics.
        \item The authors should make sure to preserve anonymity (e.g., if there is a special consideration due to laws or regulations in their jurisdiction).
    \end{itemize}

\item {\bf Broader impacts}
    \item[] Question: Does the paper discuss both potential positive societal impacts and negative societal impacts of the work performed?
    \item[] Answer: \answerYes{} 
    \item[] Justification: \autoref{discussion} discusses positive societal impacts through more efficient and accessible LLM inference, as well as negative impacts including lowered barriers for harmful or sensitive applications and safety, privacy, fairness, and robustness risks under distribution shift or poor calibration.
    \item[] Guidelines:
    \begin{itemize}
        \item The answer \answerNA{} means that there is no societal impact of the work performed.
        \item If the authors answer \answerNA{} or \answerNo, they should explain why their work has no societal impact or why the paper does not address societal impact.
        \item Examples of negative societal impacts include potential malicious or unintended uses (e.g., disinformation, generating fake profiles, surveillance), fairness considerations (e.g., deployment of technologies that could make decisions that unfairly impact specific groups), privacy considerations, and security considerations.
        \item The conference expects that many papers will be foundational research and not tied to particular applications, let alone deployments. However, if there is a direct path to any negative applications, the authors should point it out. For example, it is legitimate to point out that an improvement in the quality of generative models could be used to generate Deepfakes for disinformation. On the other hand, it is not needed to point out that a generic algorithm for optimizing neural networks could enable people to train models that generate Deepfakes faster.
        \item The authors should consider possible harms that could arise when the technology is being used as intended and functioning correctly, harms that could arise when the technology is being used as intended but gives incorrect results, and harms following from (intentional or unintentional) misuse of the technology.
        \item If there are negative societal impacts, the authors could also discuss possible mitigation strategies (e.g., gated release of models, providing defenses in addition to attacks, mechanisms for monitoring misuse, mechanisms to monitor how a system learns from feedback over time, improving the efficiency and accessibility of ML).
    \end{itemize}
    
\item {\bf Safeguards}
    \item[] Question: Does the paper describe safeguards that have been put in place for responsible release of data or models that have a high risk for misuse (e.g., pre-trained language models, image generators, or scraped datasets)?
    \item[] Answer: \answerNA{} 
    \item[] Justification: The paper does not release a new high-risk model or scraped dataset; it releases a reference implementation for a training-free quantization method and evaluates existing open-weight LLMs on public benchmarks. It also notes that any deployment of compressed LLMs should be evaluated for safety, privacy, fairness, and robustness in the target domain.
    \item[] Guidelines:
    \begin{itemize}
        \item The answer \answerNA{} means that the paper poses no such risks.
        \item Released models that have a high risk for misuse or dual-use should be released with necessary safeguards to allow for controlled use of the model, for example by requiring that users adhere to usage guidelines or restrictions to access the model or implementing safety filters. 
        \item Datasets that have been scraped from the Internet could pose safety risks. The authors should describe how they avoided releasing unsafe images.
        \item We recognize that providing effective safeguards is challenging, and many papers do not require this, but we encourage authors to take this into account and make a best faith effort.
    \end{itemize}

\item {\bf Licenses for existing assets}
    \item[] Question: Are the creators or original owners of assets (e.g., code, data, models), used in the paper, properly credited and are the license and terms of use explicitly mentioned and properly respected?
    \item[] Answer: \answerYes{} 
    \item[] Justification: The appendix (\autoref{subsec:assets_licence}) lists all existing datasets, model checkpoints, and baseline/code assets used in the experiments, including original creators, versions/checkpoints or splits, URLs, licenses, and terms-of-use notes. We use these assets only for research/evaluation, comply with their licenses, and do not redistribute third-party datasets or model weights.
    \item[] Guidelines:
    \begin{itemize}
        \item The answer \answerNA{} means that the paper does not use existing assets.
        \item The authors should cite the original paper that produced the code package or dataset.
        \item The authors should state which version of the asset is used and, if possible, include a URL.
        \item The name of the license (e.g., CC-BY 4.0) should be included for each asset.
        \item For scraped data from a particular source (e.g., website), the copyright and terms of service of that source should be provided.
        \item If assets are released, the license, copyright information, and terms of use in the package should be provided. For popular datasets, \url{paperswithcode.com/datasets} has curated licenses for some datasets. Their licensing guide can help determine the license of a dataset.
        \item For existing datasets that are re-packaged, both the original license and the license of the derived asset (if it has changed) should be provided.
        \item If this information is not available online, the authors are encouraged to reach out to the asset's creators.
    \end{itemize}

\item {\bf New assets}
    \item[] Question: Are new assets introduced in the paper well documented and is the documentation provided alongside the assets?
    \item[] Answer: \answerYes{} 
    \item[] Justification: The only new asset released is the anonymized TAQ reference implementation, provided in the abstract, and the paper documents the relevant public datasets, models, quantization protocol, splits, and hyperparameters needed to use it; no new datasets, models, or human-derived assets are introduced.
    \item[] Guidelines:
    \begin{itemize}
        \item The answer \answerNA{} means that the paper does not release new assets.
        \item Researchers should communicate the details of the dataset\slash code\slash model as part of their submissions via structured templates. This includes details about training, license, limitations, etc. 
        \item The paper should discuss whether and how consent was obtained from people whose asset is used.
        \item At submission time, remember to anonymize your assets (if applicable). You can either create an anonymized URL or include an anonymized zip file.
    \end{itemize}

\item {\bf Crowdsourcing and research with human subjects}
    \item[] Question: For crowdsourcing experiments and research with human subjects, does the paper include the full text of instructions given to participants and screenshots, if applicable, as well as details about compensation (if any)? 
    \item[] Answer: \answerNA{} 
    \item[] Justification: The paper does not involve crowdsourcing or human-subject experiments; it evaluates existing public benchmarks and open-weight LLMs, with no new sensitive-data collection or participant instructions/compensation to report.
    \item[] Guidelines:
    \begin{itemize}
        \item The answer \answerNA{} means that the paper does not involve crowdsourcing nor research with human subjects.
        \item Including this information in the supplemental material is fine, but if the main contribution of the paper involves human subjects, then as much detail as possible should be included in the main paper. 
        \item According to the NeurIPS Code of Ethics, workers involved in data collection, curation, or other labor should be paid at least the minimum wage in the country of the data collector. 
    \end{itemize}

\item {\bf Institutional review board (IRB) approvals or equivalent for research with human subjects}
    \item[] Question: Does the paper describe potential risks incurred by study participants, whether such risks were disclosed to the subjects, and whether Institutional Review Board (IRB) approvals (or an equivalent approval/review based on the requirements of your country or institution) were obtained?
    \item[] Answer: \answerNA{} 
    \item[] Justification: The paper does not involve crowdsourcing or research with human subjects; it evaluates LLM quantization using existing benchmarks and open-weight models, and states that there are no human-subject experiments or new sensitive-data collection.
    \item[] Guidelines:
    \begin{itemize}
        \item The answer \answerNA{} means that the paper does not involve crowdsourcing nor research with human subjects.
        \item Depending on the country in which research is conducted, IRB approval (or equivalent) may be required for any human subjects research. If you obtained IRB approval, you should clearly state this in the paper. 
        \item We recognize that the procedures for this may vary significantly between institutions and locations, and we expect authors to adhere to the NeurIPS Code of Ethics and the guidelines for their institution. 
        \item For initial submissions, do not include any information that would break anonymity (if applicable), such as the institution conducting the review.
    \end{itemize}

\item {\bf Declaration of LLM usage}
    \item[] Question: Does the paper describe the usage of LLMs if it is an important, original, or non-standard component of the core methods in this research? Note that if the LLM is used only for writing, editing, or formatting purposes and does \emph{not} impact the core methodology, scientific rigor, or originality of the research, declaration is not required.
    \item[] Answer: \answerYes{} 
    \item[] Justification: LLMs are central to the method: TAQ operates on frozen LLM hidden representations and output sensitivities to allocate per-layer quantization precision, and the paper describes the relevant LLM backbones, calibration prompts, and scoring/allocation procedure.
    \item[] Guidelines:
    \begin{itemize}
        \item The answer \answerNA{} means that the core method development in this research does not involve LLMs as any important, original, or non-standard components.
        \item Please refer to our LLM policy in the NeurIPS handbook for what should or should not be described.
    \end{itemize}

\end{enumerate}